%% file: main.tex
\documentclass{article}
\usepackage[utf8]{inputenc}
\PassOptionsToPackage{round,authoryear}{natbib}
\usepackage{main}
\usepackage{microtype}
\usepackage{graphicx}
\usepackage{times}
\usepackage{amsmath}
\usepackage{amssymb}
\usepackage{enumitem}
\usepackage{booktabs}
\usepackage{multirow}
\usepackage{xspace}
\usepackage{float}
\definecolor{mydarkblue}{rgb}{0,0.08,0.45}
\usepackage[colorlinks=true,linkcolor=mydarkblue,citecolor=mydarkblue,filecolor=mydarkblue,urlcolor=mydarkblue]{hyperref}
\usepackage{fancyhdr}

\setlist[itemize,1]{leftmargin=10pt}

\input{commands}

\title{\fontsize{15}{17}\selectfont Rethinking Reasoning-Intensive Retrieval:\\Evaluating and Advancing Retrievers in Agentic Search Systems}

\definecolor{YaleBlue}{RGB}{0, 53, 107}
\definecolor{NUSOrange}{RGB}{239, 124, 0}
\definecolor{NYUViolet}{RGB}{87, 6, 140}

\newcommand{\Yale}{\hspace{.1em}^{\textcolor{YaleBlue}{\boldsymbol{Y}}}}
\newcommand{\NUS}{\hspace{.1em}^{\textcolor{NUSOrange}{\boldsymbol{S}}}}
\newcommand{\NYUSH}{\hspace{.1em}^{\textcolor{NYUViolet}{\boldsymbol{N}}}}

\author{
\textbf{Yilun Zhao}$\Yale$ \qquad
\textbf{Jinbiao Wei}$\Yale$ \qquad
\textbf{Tingyu Song}$\Yale$ \qquad
\textbf{Siyue Zhang}$\NUS$ \\ [3pt]
\textbf{Chen Zhao}$\NYUSH$ \qquad
\textbf{Arman Cohan}$\Yale$ \\ [7pt]
$\Yale$Yale NLP Lab \qquad
$\NUS$National University of Singapore \qquad
$\NYUSH$NYU Shanghai \\ [6pt]
\huggingface \href{https://huggingface.co/collections/yale-nlp/rtriever}{\model}
\hspace{3em}
\github \href{https://github.com/yale-nlp/Bright-Pro}{\ours}
}

\begin{document}

\fancyhf{}
\renewcommand{\headrulewidth}{0pt}
\renewcommand{\footrulewidth}{0pt}
\setlength{\headheight}{20pt}
\setlength{\headsep}{3pt}
\lhead{%
    \raisebox{0.09cm}{\includegraphics[height=0.5cm]{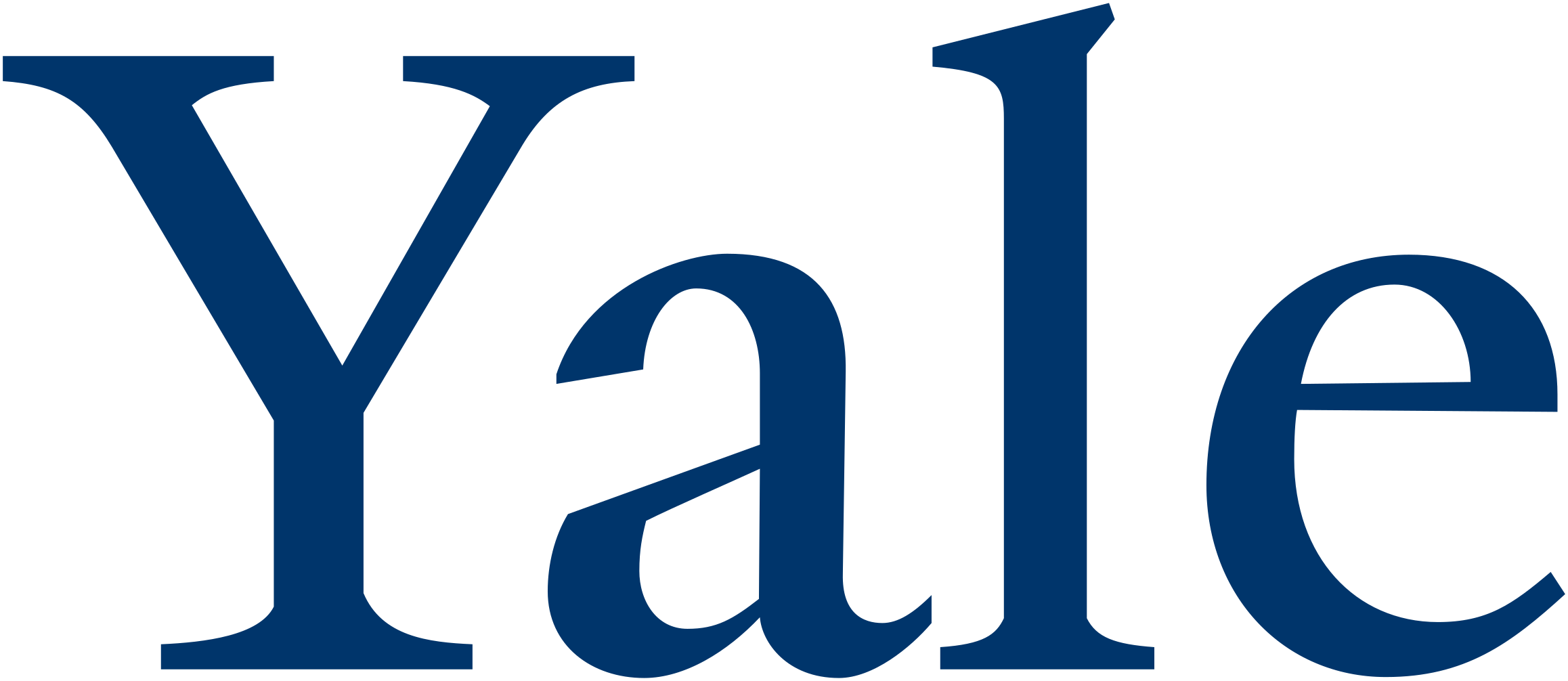}}\hspace{0.25cm}%
    \includegraphics[height=0.67cm]{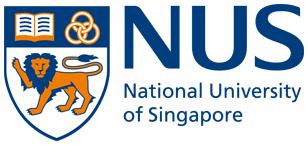}\hspace{0.25cm}%
    \includegraphics[height=0.67cm]{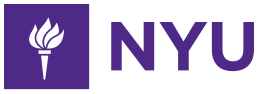}%
}
\makeatletter
\renewcommand{\@toptitlebar}{%
  \vskip -0.35cm
  {\color{black}\hrule height 1\p@}
  \vskip 0.20in
  \vskip -\parskip%
}
\makeatother

\maketitle
\thispagestyle{fancy}
\pagestyle{plain}

\begin{abstract}
\input{main/0-abstract}
\end{abstract}

\input{main/1-introduction}
\input{main/2-related_work}

\input{main/3-data}
\input{main/4-method}
\input{main/5-experiment}
\input{main/6-conclusion}

\input{main/limitations}

\bibliographystyle{unsrtnat}
\bibliography{anthology,custom}

\newpage
\appendix
\input{appendix/main}

\end{document}

%% file: commands.tex
\usepackage{tikz}
\usepackage{soul}
\usepackage{colortbl}
\usepackage{subcaption}
\usepackage{bbm}
\usepackage{adjustbox}
\usepackage{makecell}
\usepackage{tabularx}
\usepackage{tabu}
\usepackage{pifont}
\usepackage{algorithmicx}
\usepackage{algorithm}
\usepackage{algpseudocode}
\usepackage{arydshln}
\usepackage{bm}
\usepackage{dsfont}
\usepackage{tablefootnote}
\usepackage[most]{tcolorbox}
\usepackage{wrapfig}

\newcommand{\eg}{\hbox{\emph{e.g.,}}\xspace}
\newcommand{\ie}{\hbox{\emph{i.e.,}}\xspace}

\newcommand{\cmark}{\ding{51}}
\newcommand{\xmark}{\ding{55}}

\newcommand{\ind}[1]{\mathds{1}\{#1\}}

\newcolumntype{C}[1]{>{\centering\arraybackslash}m{#1}}
\newcolumntype{L}[1]{>{\raggedright\arraybackslash}m{#1}}

\tcbset{
  colback=black!4,
  colframe=black!60,
  boxsep=1.2mm,
  left=1.2mm,right=1.2mm,top=1.2mm,bottom=1.2mm,
  title filled=true
}

\newcommand{\ours}{\textsc{Bright-Pro}\xspace}
\newcommand{\model}{RTriever\xspace}
\newcommand{\bright}{\textsc{Bright}\xspace}
\newcommand{\instructor}{\textsc{InstructOR}\xspace}
\newcommand{\andcg}{$\alpha$-nDCG@$k$\xspace}
\newcommand{\andcgb}{$\bm{\alpha}$-nDCG@$\mathbf{k}$\xspace}

\newcommand{\huggingface}{\raisebox{-1.5pt}{\includegraphics[height=1.05em]{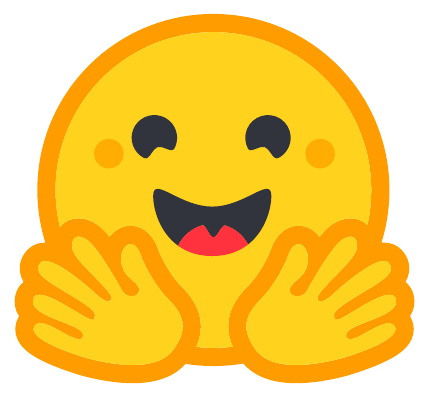}}\xspace}
\newcommand{\github}{\raisebox{-1.5pt}{\includegraphics[height=1.05em]{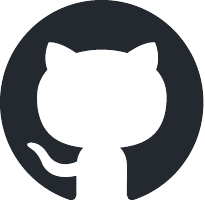}}\xspace}

%% file: main/0-abstract.tex
Reasoning-intensive retrieval aims to surface evidence that supports downstream reasoning rather than merely matching topical similarity.
This capability is increasingly important for agentic search systems, where retrievers must provide complementary evidence across iterative search and synthesis.
However, existing work remains limited on both evaluation and training: benchmarks such as \bright provide narrow gold sets and evaluate retrievers in isolation, while synthetic training corpora often optimize single-passage relevance rather than evidence portfolio construction.
We introduce \ours, an expert-annotated benchmark that expands each query with multi-aspect gold evidence and evaluates retrievers under both static and agentic search protocols.
We further construct \model-Synth, an aspect-decomposed synthetic corpus that generates complementary positives and positive-conditioned hard negatives, and use it to LoRA fine-tune \model-4B from Qwen3-Embedding-4B.
Experiments across lexical, general-purpose, and reasoning-intensive retrievers show that aspect-aware and agentic evaluation expose behaviors hidden by standard metrics, while \model-4B substantially improves over its base model.

%% file: main/1-introduction.tex
\input{figures_tex/figure1}

\section{Introduction}
Information Retrieval (IR) has long served as the foundation for accessing and organizing knowledge at scale~\cite{scholarqa, litsearch, beir}. Traditional IR systems have achieved remarkable success in scenarios where user intent can be satisfied by retrieving factoid or single-hop information~\cite{qwen3embed, promptriever, nv-embed}. 
However, as user queries become increasingly complex and demand multi-step reasoning, retrieval systems must move beyond surface-level relevance, giving rise to \emph{reasoning-intensive retrieval}, a new frontier in IR research~\cite{rar-b, su2025bright}.

In practice, reasoning-intensive queries often require multi-step reasoning and the integration of diverse evidence, which causes retrievers to struggle to surface sufficiently relevant information within a single retrieval step. 
To address this limitation, recent work has turned to complex, multi-stage information-seeking pipelines, culminating in the emergence of \emph{Deep-Research}~\cite{chen2025browsecomp, deepresearchbench, swesmith, webwalker, zhao2026sciarena}. Deep-research systems employ LLM-based agents that iteratively plan, search, read, and synthesize information in order to answer difficult queries, largely compensating for the limited effectiveness of integrated retrievers.
Because relevant evidence is rarely surfaced in a single retrieval step, these systems rely on increased iterations of retrieval and reasoning, leading to substantial computational cost and latency.
This reliance highlights the importance of reasoning-intensive retrieval: a retriever that surfaces high-quality, reasoning-ready evidence in a single step would reduce iteration cost and improve deep-research efficiency.

Despite growing interest in reasoning-intensive retrieval, both evaluation and training remain misaligned with this agentic use case. On the evaluation side, the main existing benchmark, \bright~\cite{su2025bright}, provides each query with only a small set of gold passages, typically derived from one or two web pages, and evaluates retrievers in isolation rather than inside a dynamic deep-research loop. On the training side, recent synthetic corpora for reasoning retrieval~\cite{shao2025reasonir,long2025diver} often pair each query with a single positive and hard negatives, encouraging models to rank \emph{a} relevant passage highly rather than to retrieve complementary evidence that jointly supports complex reasoning. As a result, a retriever can appear strong under single-passage metrics while still failing to cover the reasoning aspects needed by an agent.

To bridge the evaluation gap, we introduce \textbf{\ours}, an evaluation framework that extends \bright with richer, multi-aspect supervision for reasoning-intensive retrieval. Expert annotators expand each query's gold passage set and group passages into \emph{reasoning aspects} that reflect distinct perspectives or subproblems within the same query, as illustrated in \autoref{fig:example}. This enables fine-grained analysis of whether a retriever covers the full reasoning need rather than concentrating on a single aspect. Beyond static evaluation, we integrate retrievers into an LLM-based agentic workflow that iteratively plans, retrieves, and synthesizes information, allowing us to measure system-level outcomes such as reasoning completeness, iteration efficiency, and final response quality.

To address the training gap, we construct \model-Synth, an aspect-decomposed synthetic corpus that teaches retrievers complementary evidence selection. Starting from MS MARCO seeds, our pipeline rewrites short queries into realistic analytical ones, generates a self-contained reference answer, decomposes it into non-overlapping reasoning aspects, and realizes each aspect as a positive passage. It then generates positive-conditioned hard negatives: passages that share topical cues but deliberately omit the needed aspect. We use this corpus to LoRA fine-tune \textbf{\model-4B} from Qwen3-Embedding-4B~\cite{qwen3embedding}.

We evaluate classical lexical, general-purpose, and reasoning-intensive retrievers on \ours. Across subsets and settings, BGE-Reasoner~\cite{bge_reasoner} consistently delivers the strongest performance, while the DIVER family~\cite{long2025diver} forms a strong second tier and Qwen-family embedders~\cite{gte,qwen3embedding} trail by a noticeable gap. Our \model-4B lifts the 4B base by a wide margin and lands in the upper-middle tier across settings, ahead of much larger general-purpose embedders. Although BM25 performs poorly in static retrieval, it becomes competitive under the agentic protocol, where targeted follow-up queries reduce vocabulary mismatch and allow lexical matching to surface useful evidence. Case studies further show that early access to gold evidence shortens interaction traces and improves reasoning completeness, whereas weaker retrievers invite speculative reasoning and recycle low-utility passages. Reasoning-intensive retrieval thus calls for joint progress on evaluation and training to support iterative research workflows.

Our main contributions are summarized below:
\begin{itemize}[leftmargin=*]
\itemsep0em 
\item We introduce \ours, an expert-annotated benchmark that extends \bright with multi-aspect evidence and evaluates retrievers under both static and agentic search settings.

\item We propose \model-Synth, an aspect-decomposed synthesis pipeline that generates complementary positives from reference-answer reasoning and positive-conditioned hard negatives for training retrievers.

\item We train \model-4B by LoRA fine-tuning Qwen3-Embedding-4B on \model-Synth, providing a 4B retriever specialized for reasoning-intensive evidence selection.

\item We deliver an empirical study that disentangles retriever quality from agent-retriever fit, showing that aspect-aware and agentic protocols expose retrieval behaviors hidden by standard rankings.
\end{itemize}

%% file: figures_tex/figure1.tex
\begin{figure}[h]
    \centering
    \includegraphics[width = 0.93\linewidth]{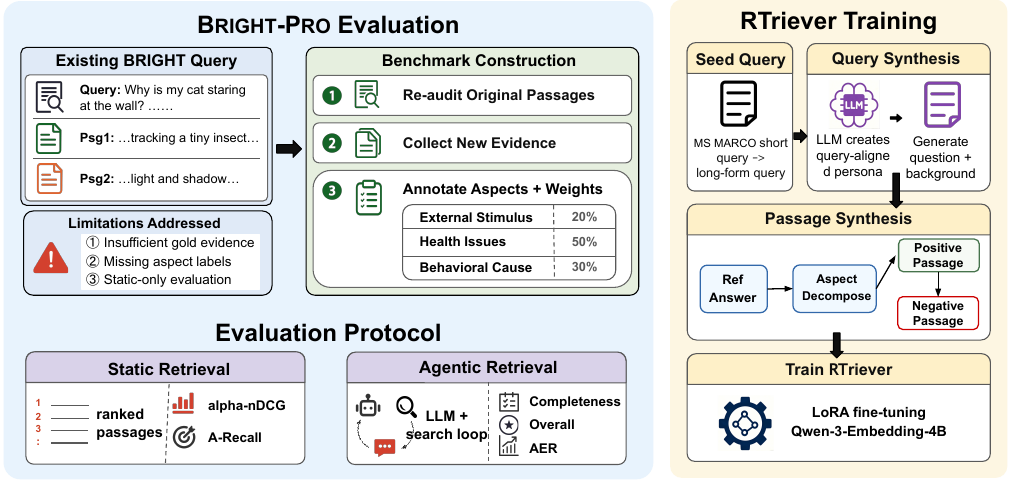}
    \caption{
        Overview of our work. \emph{Left:} \ours augments \bright with re-audited gold passages and reasoning-aspect-level labels, enabling retriever evaluation under both static and agentic search protocols. \emph{Right:} \model is trained on \model-Synth. \model-Synth rewrites MS MARCO queries into DeepResearch-style queries, generates reference answers and decomposes them into non-overlapping reasoning aspects, then synthesizes complementary positives for each aspect along with positive-conditioned hard negatives for LoRA fine-tuning.
    }
    \label{fig:example}
\end{figure}

%% file: main/2-related_work.tex
\input{figures_tex/data_pipeline}
\section{Related Work}
\paragraph{Reasoning-Intensive Information Retrieval.}
Beyond traditional keyword- and semantic-based IR, \bright~\cite{su2025bright} is the first benchmark that explicitly targets queries requiring multi-step reasoning to identify genuinely useful evidence rather than merely superficial relevance, with concurrent extensions to multimodal~\cite{Abdallah2026MMBRIGHTAM,Zhang2025MRMRAR} and instruction-following~\cite{Song2025IFIRAC} settings. Building on this formulation, recent work has trained retrievers on synthetic data containing reasoning-intensive queries and hard negatives to encourage reasoning-aware evidence selection~\cite{weller2025rank1testtimecomputereranking, das2025rader, zhang2025diffusion, long2025diver, shao2025reasonir, Jin2026LaSERIE, Chen2026AgentIRRR, Song2025LimRankLI}. However, current evaluations typically score a retriever against a small pool of gold passages tied to one or two sources~\cite{Abdallah2026AreLR}, which limits assessment of coverage across complementary reasoning aspects.

\paragraph{Agentic Search System and Evaluation.} In parallel, agentic search systems (\eg DeepResearch) combine LLM planning with iterative search, reading, and synthesis to tackle complex queries~\cite{chen2025browsecomp, swesmith, webwalker, shao2025drtulu, Ding2026SciRAGAC}.
Several benchmarks are proposed to evaluate the DeepResearch system responses~\cite{wang2025liveresearchbench, deepresearchbench, yifei2025researchqa, Xiong2026AutoResearchBenchBA, Li2026DeepResearchBI, Gupta2026DeepSearchQABT, Wu2026AgentSearchBenchAB, Hu2026SAGEBA}.
To enable controlled comparisons of agent components, \textsc{BrowseComp-Plus}~\cite{chen2025browsecomp} provides a fixed, curated corpus that standardizes access and reduces environmental variance. While this improves reproducibility, it abstracts away open-domain retrieval dynamics and offers limited visibility into how a retriever shapes an agent's evidence portfolio, iteration budget, and final reasoning quality, gaps echoed by recent calls for explicit information-coverage evaluation~\cite{Samuel2026CoverageBenchEI}.

%% file: figures_tex/data_pipeline.tex
\begin{figure*}[!t]
 \centering
\includegraphics[width=\textwidth]{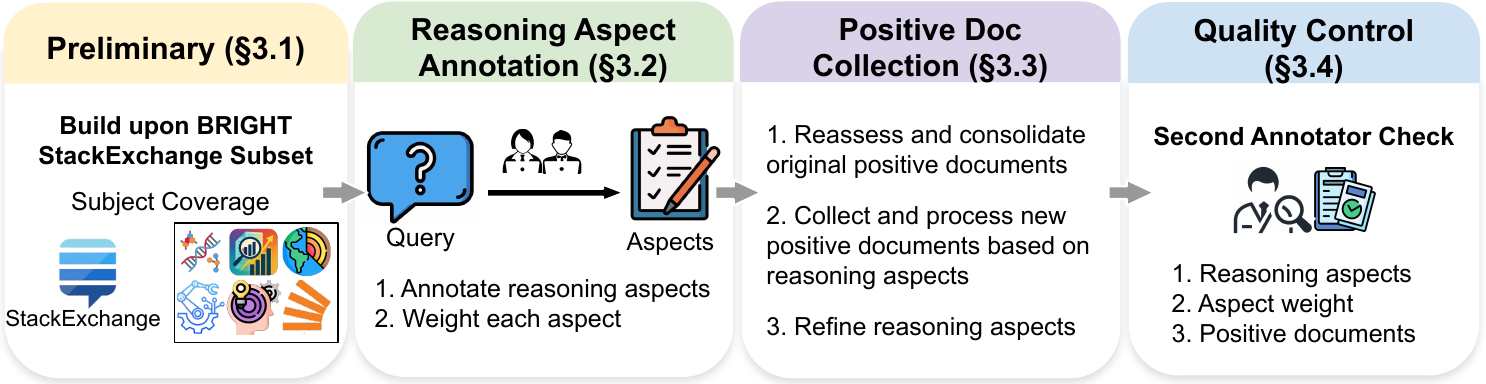}
 \caption{An overview of the \ours benchmark construction pipeline.}
 \label{fig:data-pipeline}
\end{figure*}

%% file: main/3-data.tex
\section{\ours Benchmark}
This section introduces \ours, which builds upon \bright~\cite{su2025bright} to enable a more thorough evaluation of reasoning-intensive retrieval.
\autoref{fig:data-pipeline} outlines the construction pipeline.

\subsection{Preliminary: \bright Benchmark}
The \bright benchmark 
is divided into three subsets: (1) \emph{StackExchange},  which focuses on retrieving passages that support answering questions sourced from StackExchange; (2) \emph{Coding}, which focuses on retrieving documentation or analogous solved problems; and (3) \emph{Theorem}, which targets retrieving solved problems that use the same theorems or contain relevant theorem statements.
\ours builds upon the StackExchange subset because it best represents open-domain, natural language reasoning. In contrast, the coding and theorem subsets rely heavily on domain-specific syntax or formal logic, making them less suitable for evaluating general-purpose reasoning retrievers.
For queries in the StackExchange subset, annotators trace hyperlinks in accepted or high-quality answers to collect the specific web pages that those answers depend on. Each web page is then segmented into multiple passages, which serve as gold passages. To construct negative samples, annotators use Google Search with either the post title or LLM-generated keywords to locate topically related web pages that do not meet the precise informational need of the query. 
All collected passages are then compiled into a unified retrieval corpus.

\subsection{Reasoning Aspect Annotation}\label{sec:data-aspect}
In real-world reasoning-intensive retrieval scenarios, users seek a comprehensive set of evidence that collectively supports the entire reasoning chain behind a query. Reasoning completeness often decomposes into multiple reasoning aspects, where each aspect represents a coherent subproblem or perspective. When a retriever exhibits bias toward passages aligned with only one aspect, the resulting evidence portfolio becomes unbalanced, thereby weakening downstream synthesis. 

\paragraph{Reasoning Aspect Annotation.}
Motivated by this, for each query from the \bright StackExchange subset we assign field-specific expert annotators and ask them to decompose the information need into a compact set of \emph{reasoning aspects}. We provide links to community answers on StackExchange as a starting point. Because user-contributed answers can be concise, audience-tailored, heuristic, or outdated, annotators must think beyond them and independently identify the premises needed for a defensible solution, each documented with a one- to two-sentence rationale.

\paragraph{Aspect Weight Annotation.}
To reflect unequal contributions across aspects, we explicitly weight aspects rather than treating them as interchangeable. This design focuses evaluation on what most affects correctness, reduces gaming by accumulating low-value snippets, and aligns scoring with the utility of the final synthesis. For each reasoning aspect, annotators assign Likert scores from 1 to 5 based on the aspect’s importance to producing a correct and defensible final answer (the detailed rubric is provided in Appendix~\ref{app:rebric}).
We then nomalize the scores to weights $w_a \in [0,1]$ with $\sum_a w_a = 1$.

\subsection{Aspect-Guided Positive Doc Collection}
After establishing the reasoning aspects and their weights, we proceed to collect positive documents for each query. 

\paragraph{Reassessment and Consolidation of Original Positive Documents.}
The \bright benchmark provides a set of positive passages. However, these passages are not assigned to reasoning aspects and, in our preliminary screening, a nontrivial portion was weakly related to the underlying queries. Therefore, we first ask annotators to re-audit the original \bright positives: each passage is reviewed for topical fidelity to the query, and either assigned to current aspects or discarded if it fails to provide explicit, verifiable evidence. 
Because many \bright positives were split from the same source page, we further reconcile passages that are contiguous or substantively overlapping. When multiple passages from a single URL support the same aspect, annotators merge them into one coherent segment to preserve context and reduce redundancy, while trimming unrelated text. Each merged segment is stored as a single positive tied to its aspect.

\paragraph{Acquisition and Processing of New Positive Documents.}
Annotators are allowed to use both conventional web search or AI-assisted search like Perplexity AI and ChatGPT with Web Search when searching for aspect-relevant documents. 
A document is accepted as a positive instance only if it provides explicit, precise, and credible evidence supporting one or more aspects. 
We provide annotators with a customized interface built upon the FireCrawl framework. Annotators input the URL of a web page, and the interface automatically downloads the HTML source, removes boilerplate content (\eg advertisements, navigation menus), and extracts the textual content for further review.
Annotators are then required to manually refine the extracted text to remove residual noise or irrelevant segments. 
If a document’s content covers multiple reasoning aspects, annotators must segment the text into aspect-specific portions so that each segment aligns with a single aspect. Each segmented portion is then stored as an independent positive sample associated with its respective aspect.

\paragraph{Iterative Refinement of Reasoning Aspects.}
Document collection is an iterative process that naturally leads to refinement of the aspect structure. As annotators search, they often uncover missing assumptions, dependencies between aspects, or cases where two aspects capture overlapping reasoning needs. In such cases, annotators carefully adjust the aspect schema, \eg, by clarifying definitions to eliminate redundancy, subdividing overly broad aspects into more precise ones, or consolidating highly correlated aspects that contribute to the same reasoning process.

\input{tables/data_statistics}
\subsection{Data Quality Control}
Each example in \ours is independently re-examined by a second annotator from the same field, who verifies both the reasoning aspects (their granularity, balance, weights, and coverage of the reasoning process) and the supporting documents (whether each offers explicit, credible evidence for its assigned aspect). To assess weighting reliability, we randomly sample 50 queries and ask independent reviewers to rescore the annotated reasoning aspects; the resulting weighted Cohen's $\kappa$ of $0.742$ indicates stable importance ratings across annotators. The full review protocol is described in Appendix~\ref{app:data-quality}.
\autoref{tab:data-stat} presents the key statistics of the \ours benchmark. Examples for each subset can be found in Appendix~\ref{app:data-example}.

\section{\ours Evaluation Protocol}
We evaluate retrievers along two complementary dimensions: a \emph{static} setting, which isolates retrieval quality using aspect-annotated gold passages, and an \emph{agentic} setting, which assesses the retriever inside a deep-research workflow.

\subsection{Static Retrieval Evaluation}
We first evaluate retrievers in a traditional IR setting, where each query receives a single ranked list of passages. Because every query is annotated with reasoning aspects of varying importance (Section~\ref{sec:data-aspect}), we adopt a metric that rewards covering complementary aspects rather than over-retrieving from a single one. Our primary metric is \textbf{\andcg}~\cite{andcg} with novelty penalty $\alpha=0.5$, which discounts repeated retrieval from already-covered aspects. We complement it with Weighted Aspect Recall (\textbf{A-Recall@$k$}), which credits each aspect once covered, and report \textbf{NDCG@$k$} and \textbf{Recall@$k$} as diagnostics. See Appendix~\ref{app:static-metrics} for full definitions.

\subsection{Agentic Pipeline and Response Evaluation}
\label{sec:agentic-pipeline}
To measure retrievers as components of deep-research systems, we plug each retriever into the same LLM agent equipped with a single search tool. The retriever is the only experimental variable: the agent receives the original query, iteratively issues search queries over the corresponding corpus, reads the returned passages, and produces a citation-grounded final answer. We run this protocol with two agent backends, GPT-5-mini-08-07 and Qwen3.5-122B-A10B~\cite{yang2025qwen3technicalreport}, using identical prompts, tool interfaces, retrieval depth, and stopping budgets; implementation details are provided in Appendix~\ref{app:agent-config}, and prompt templates in Appendix~\ref{app:prompt}.

We evaluate final answers with GPT-5 as an LLM-as-Judge, following prior work on deep-research evaluation~\cite{chen2025browsecomp,li2024llmsasjudges}. The judge compares each system answer against a reference answer generated from the human-annotated reasoning aspects and their positive passages. It assigns each aspect a coverage score in $\{0,0.5,1\}$ and an answer-level overall quality score from 1 to 5. We aggregate aspect coverage using the annotated aspect weights and report it as reasoning completeness, together with overall quality. Reference construction and validation are detailed in Appendix~\ref{app:ref-answer-validation}; the judge prompt appears in Figure~\ref{fig:judge-prompt}.

\subsection{Fixed- and Adaptive-Round Protocols}
\label{sec:agentic-protocols}
To bound cost while preserving comparability, all agentic experiments use the same fixed sample of 175 queries randomly sampled from \ours, with 25 queries per task.

\paragraph{Fixed-Round Protocol.}
The agent performs exactly $R \in \{1,2,3\}$ search rounds. Each round returns the top-$5$ passages, and after each round the agent generates an answer conditioned on all evidence retrieved so far. This protocol isolates retriever quality under matched interaction budgets. We report cumulative \textbf{\andcgb} at $5R$, with reasoning completeness and overall answer quality.

\paragraph{Adaptive-Round Protocol.}
The agent decides when to stop searching based on whether the accumulated evidence is sufficient. This protocol measures both answer quality and interaction efficiency: a stronger retriever should satisfy the agent's evolving information need in fewer rounds. We report the number of retrieval rounds, reasoning completeness, overall quality, and the \textbf{Efficiency-Quality Reward (AER)}:
\begin{equation}
\mathrm{AER} = OQ \times e^{-\gamma (R - 1)}
\end{equation}
where \(OQ\) is the overall quality of the response, \(R\) is the number of search rounds, and $\gamma=0.05$.

%% file: tables/data_statistics.tex
\begin{table}[!t]
\centering
\small
\begin{tabular}{l|rrrr|rrr|c}
\toprule
\multirow{2}{3.5em}{\centering \bf{Subset}} & \multicolumn{4}{c|}{\textbf{Total Number}} & \multicolumn{3}{c|}{\textbf{Avg. Length}} & \multirow{2}{3.5em}{\centering \bf{Example}} \\
\cmidrule{2-5}\cmidrule{6-8}
& \multicolumn{1}{c}{$\mathbf{Q}$} & \multicolumn{1}{c}{$\boldsymbol{\mathcal{D}}$} & \multicolumn{1}{r}{$\boldsymbol{\mathcal{D}^+}$} & \multicolumn{1}{c}{$\boldsymbol{\mathcal{A}}$} & \multicolumn{1}{c}{$\mathbf{Q}$} & \multicolumn{1}{c}{$\boldsymbol{\mathcal{D}}$} & \multicolumn{1}{c}{$\boldsymbol{\mathcal{A}}$} \\
\midrule
Biology            & 103 & 59{,}513  & 7.81  & 3.94 & 92.6  & 75.7  & 20.1 & Tab.~\ref{tab:biology-example} \\
Earth Science      & 115 & 123{,}575 & 7.44  & 3.83 & 82.2  & 41.8  & 20.1 & Tab.~\ref{tab:earth-science-example} \\
Economics          & 99  & 52{,}240  & 7.81  & 3.71 & 123.5 & 80.4  & 19.9 & Tab.~\ref{tab:economics-example} \\
Psychology         & 100 & 54{,}741  & 7.07  & 3.84 & 116.2 & 72.7  & 19.8 & Tab.~\ref{tab:psychology-example} \\
Robotics           & 101 & 63{,}920  & 6.17  & 3.71 & 218.8 & 45.6  & 19.1 & Tab.~\ref{tab:robotics-example} \\
Stack Overflow     & 115 & 109{,}188 & 4.60  & 3.32 & 172.0 & 151.8 & 19.0 & Tab.~\ref{tab:stackoverflow-example} \\
Sustain. Living    & 106 & 63{,}142  & 9.25  & 3.86 & 116.9 & 69.2  & 20.2 & Tab.~\ref{tab:sustainable-living-example} \\
\midrule
Overall            & 739 & 526{,}319 & 7.13  & 3.74 & 131.4 & 79.3  & 19.7 & --- \\
\bottomrule
\end{tabular}
\caption{
Data statistics of \ours.
For each subset, we show the number of queries ($\mathbf{Q}$) and documents ($\boldsymbol{\mathcal{D}}$), the average number of positive documents per query ($\boldsymbol{\mathcal{D}^+}$) and the average number of reasoning aspects per query ($\boldsymbol{\mathcal{A}}$), together with the average word counts of queries, documents, and aspect descriptions.
}
\label{tab:data-stat}
\end{table}

%% file: main/4-method.tex
\section{\model: Learning to Retrieve Complementary Evidence}
\label{sec:method}

Reasoning-intensive retrieval is rarely consumed in isolation. Inside a deep-research agent, the retriever is queried repeatedly across rounds, and the value of any single retrieval is judged by whether the accumulated set of passages \emph{collectively} satisfies the user's reasoning need. Yet current synthetic corpora for reasoning-intensive retrieval~\cite{shao2025reasonir,long2025diver} pair every query with a single synthetic positive, so a contrastive retriever trained on them learns to rank \emph{a} relevant document highest, not to surface a balanced portfolio of complementary evidence. We close this gap with \textbf{\model-Synth}, a synthetic corpus whose every analytical query is paired with a set of complementary positives that together span the reasoning aspects of the query.

\subsection{Synthesis Pipeline}
\label{sec:method-pipeline}
The pipeline runs in two stages: it first rewrites each MS MARCO seed into a realistic deep-research query, then decomposes a reference answer into reasoning aspects, realizes each aspect as a complementary positive passage, and synthesizes matched hard negatives.

\paragraph{Realistic Query Surface from Real Seeds.} MS MARCO queries provide broad topical coverage, but many are short search-engine queries rather than the long-form information needs faced by reasoning-intensive retrievers. To convert these terse seeds into more natural query surfaces while preserving topical control, we start from MS MARCO~\cite{Campos2016MSMA} queries as semantic seeds and, for each seed, sample three personas from PersonaHub~\cite{Chan2024ScalingSD} as format examples. We then prompt an LLM to create a new persona aligned with the seed query, and to rewrite the seed query in that persona's voice as a DeepResearch-style post with a question and short background; the concatenation is what the retriever consumes. A subsequent LLM classifier labels the query as \emph{factual} or \emph{analytical}: factual queries ask for a specific piece of information that can typically be supported by a single passage, whereas analytical queries require multiple aspects to jointly support the answer.

\paragraph{Aspect-Decomposed Gold via Reference Answers.} The central design choice in our pipeline is that we do \emph{not} synthesize positives directly from the query. For each analytical query, we first prompt a strong LLM to produce a comprehensive, self-contained answer of the kind a careful expert would give once all evidence is in hand. We then ask a second LLM call to decompose the reasoning of that answer into two to three \emph{non-overlapping} aspects, each tied to an explicit rationale that names which part of the answer the aspect supplies. For each aspect, the LLM produces a passage blueprint consisting of the rationale, passage type, source, title, and a three-sentence TL;DR. A separate passage-realization call then instantiates each blueprint as a full positive passage in the indicated source style. Two properties follow by construction: every positive is load-bearing for the reasoning chain, and the positives within a query are complementary rather than restatements, so any single passage answers part of the query while the full set jointly answers it. The resulting bundles train the retriever to favor complementary evidence portfolios over isolated high-scoring passages. Factual queries skip decomposition and produce a single positive blueprint.

\paragraph{Positive-Conditioned Hard Negatives.} After the positive passage blueprints are fixed, we synthesize an equal number of hard negatives conditioned on the query and the positive passage summaries. The negative generator sees the titles and TL;DRs of the positives and is instructed to propose passages that share topical or lexical cues with the query but fail the user's actual information need when compared with those positives. For each negative, it outputs a rationale, a failure justification, source/type/title metadata, a TL;DR, and an explicit missing aspect to avoid. In the final passage-realization step, positives and negatives are generated from their blueprints using the same source- and genre-conditioned writer; for negatives, the missing aspect is added as an avoidance constraint, yielding passages that remain plausible and topically close while deliberately omitting the evidence needed to answer the query.

\subsection{\model Training Details}
\label{sec:rtriever-training}
From one million MS MARCO queries we randomly sample 140K, and the synthesis pipeline generates 140K complete (query, positives, negatives) bundles after filtering. For each training step, we pair every query with one randomly-sampled positive passage and one randomly-sampled negative passage.
We obtain \textbf{\model-4B} by LoRA fine-tuning Qwen3-Embedding-4B~\cite{qwen3embedding} on this sampled subset of \model-Synth, using the \textsc{ms-swift} framework~\cite{Zhao2024SWIFTASL}. LoRA adapters are attached to all linear projection layers of the backbone with rank $r{=}16$ and scaling factor $\alpha{=}32$, while the original embedding parameters remain frozen. Training optimizes a contrastive \textsc{InfoNCE} objective with temperature $\tau{=}0.02$ on query–document pairs: each query is paired with its positive document and one synthesized hard negative, and other documents in the same batch additionally serve as in-batch negatives. We do not include query–query or document–document contrastive pairs.
The model is trained for 5 epochs with a peak learning rate of $1{\times}10^{-5}$, a 5\% linear warm-up, and bf16 mixed-precision optimization through DeepSpeed ZeRO-2; we use a per-device batch size of 384 over 2 NVIDIA B200 GPUs (effective batch 768). Sequences are truncated to 2{,}048 tokens.

%% file: main/5-experiment.tex
\section{Experiment}

\subsection{Experiment Setup}
We benchmark \ours against twelve publicly available retrievers in addition to our \model-4B, organized into three categories:
(1) \emph{Classical Lexical Baselines}: BM25~\cite{robertson2009probabilistic}.
(2) \emph{General-Purpose Retrievers}: GritLM (7B)~\cite{muennighoff2024generative}, \instructor-XL (1.5B)~\cite{su2022one}, GTE-Qwen2-7B-Instruct (GTE-7B)~\cite{li2023towards}, Qwen3-Embedding-8B~\cite{qwen3embedding}, EmbeddingGemma-300M~\cite{Vera2025EmbeddingGemmaPA}, and OpenAI text-embedding-3-Large~\cite{neelakantan2022text}.
(3) \emph{Reasoning-Intensive Retrievers}: ReasonIR-8B~\cite{shao2025reasonir}, DIVER-Retriever-4B together with its more recent checkpoint DIVER-Retriever-4B-1020~\cite{long2025diver}, BGE-Reasoner-Embed-Qwen3-8B~\cite{bge_reasoner}, and INF-Retriever-v1-Pro (7B)~\cite{infly-ai_2025}.
For each baseline retriever, we employ its native tokenizer, embedding interface, and pooling strategy without any further fine-tuning. Input sequences are truncated to each model's validated maximum context length. For instruction-tuned retrievers, we apply either the same prompts used in \bright or model-recommended instructions where applicable to ensure fair and consistent evaluation.
All evaluation experiments are conducted on a cluster of NVIDIA H200 GPUs.

\input{tables/retriever}
\input{tables/fix-round}
\subsection{Main Findings}

\paragraph{Static Retrieval.}
Aspect-aware evaluation creates a clear separation that \bright NDCG@10 (adjacent column in \autoref{tab:model_performance}) does not: four reasoning-intensive retrievers (BGE-Reasoner-8B, DIVER-4B-1020, DIVER-4B, INF-Retriever-Pro) plus our \model-4B form a tight upper tier 4--14 points above every general-purpose embedder, including 8B-parameter Qwen3-Embedding-8B and OpenAI text-embedding-3-Large.
Training objective dominates parameter count: \model-4B (4B) beats every 7--8B general-purpose retriever, while ReasonIR-8B (although reasoning-trained) falls to eleventh of thirteen at \(\alpha\)-nDCG@25 = 41.0 because its single-positive-per-document pipeline optimizes the \bright-style single-passage signal rather than multi-aspect coverage.

\paragraph{Fixed-Round Agentic Retrieval.}
Static rankings translate only loosely to the agentic loop. BGE-Reasoner-8B leads on both axes (Round-3 \(\alpha\)-nDCG@15 = 63.0, +9.9 over the next retriever; Overall = 4.31), and DIVER-4B finishes ahead of its newer DIVER-4B-1020 sibling (Overall 4.29 vs.\ 4.16) despite trailing statically. Our \model-4B places third on Overall (4.25), close to DIVER-4B and ahead of every general-purpose retriever. Below the top tier, retrieval rank (\(\alpha\)-nDCG) and answer rank (Overall) diverge; the case studies in \S\ref{sec:qualitative} trace the mechanisms behind this divergence.
BM25 climbs from worst statically within this evaluation set (40.3) to a Round-3 \(\alpha\)-nDCG@15 of 51.5, and LLM-issued follow-ups with concrete keywords close its vocabulary-mismatch gap, a deployment-relevant pattern that static evaluations fail to surface.

\input{tables/dynamic-round}
\paragraph{Adaptive-Round Agentic Retrieval.}
The adaptive setting separates retrievers that converge quickly from those merely \emph{eventually} good.
BGE-Reasoner-8B again leads (AER = 3.65) with the fewest rounds (5.10), and our \model-4B is second by mean AER across the two agents (GPT-5-mini AER = 3.51 over 6.01 rounds); in contrast, GTE-7B has the highest overall quality (4.51) yet its 6.67 rounds drop AER to 3.44, the failure mode AER is designed to surface, since deployment cost scales with rounds, not just final answer quality.
Switching to the Qwen3.5-122B-A10B agent preserves the top tier but collapses DIVER-4B-1020 from second to seventh (AER 3.56 \(\to\) 3.11), while our \model-4B \emph{rises} to second (3.38, 0.06 behind BGE-Reasoner). Moreover, we observe that top-tier rankings remain stable, but lower-tier rankings reflect retriever-agent compatibility more than retriever quality alone.

\subsection{Qualitative Analysis}
\label{sec:qualitative}
Inspecting 175 \model-4B + GPT-5-mini adaptive-round traces surfaces five recurring patterns; we describe each below and illustrate it with a real run from the same retriever-agent stack.

\paragraph{Early-Round Retrieval Efficiency.}
When retrieval surfaces most of the gold evidence in the first one or two rounds, the agent can compose a directly grounded answer and terminates without further exploration. For example, \autoref{tab:case-early-round} shows an Antarctic ice-sheet question resolved in three rounds, with all four reasoning aspects covered.

\paragraph{Evidence Deprivation and Speculative Reasoning.}
When retrieval fails to surface any gold passage, the LLM still has to answer, so it improvises a plausible-sounding response from whatever the retriever returned, and the answer is invariably speculative. We see this in the Gazebo Garden case (\autoref{tab:case-evidence-deprivation}), which runs thirteen rounds without retrieving either gold document and incorrectly concludes the plugin is deprecated.

\paragraph{Repetition Bias and Self-Reinforcing Retrieval Errors.}
The retriever can lock onto a topic-adjacent cluster in early rounds; subsequent rounds then keep resurfacing the same passages even when the agent rephrases the query in markedly different ways, so the evidence frontier fails to expand. Consider \autoref{tab:case-repetition}, where twelve rounds yield only 28 unique documents from 60 retrieval slots, and the two highest-weighted aspects are never reached.

\paragraph{Aspect Tunnel Vision.}
On multi-aspect questions, every search round can elaborate the same aspect while other required aspects receive zero retrieval coverage. The search queries themselves remain novel each round, so the lock-in is semantic rather than document-level, but the effect on aspect-aware metrics is the same. A representative run appears in \autoref{tab:case-aspect-tunnel}: all seven rounds drill into only one half of a two-part GHCN climate question.

\paragraph{Hypothesis Hopping after Early Success.}
The converse of early termination occurs when the agent retrieves the gold passages in the first one or two rounds but cannot stop, and instead spends extra rounds testing alternative concept labels that add little. As one illustration, in \autoref{tab:case-hypothesis-hopping} round~2 already retrieves the \emph{Baader--Meinhof phenomenon} gold, yet the run continues for four more rounds, paying the deployment cost AER (\autoref{tab:iter_results}) penalises.

%% file: tables/retriever.tex
\begin{table*}[!t]
\centering
\small
\resizebox{\linewidth}{!}{
\begin{tabular}{l|C{1.1cm}|*{6}{C{1.25cm}}C{1.5cm}|C{1.1cm}}
\toprule
\textbf{Model} &
\makecell{\textbf{\bright}\\\textbf{Overall}} &
\textbf{Biology} &
\makecell{\textbf{Earth}\\\textbf{Science}} &
\textbf{Economics} &
\textbf{Psychology} &
\textbf{Robotics} &
\makecell{\textbf{Stack}\\\textbf{Overflow}} &
\makecell{\textbf{Sustainable}\\\textbf{Living}} &
\textbf{Overall} \\
\midrule
BGE-Reasoner-8B   & 33.8             & \textbf{73.5} & \textbf{74.6} & \textbf{66.0} & \textbf{65.5} & \textbf{64.7} & \textbf{67.7} & \textbf{63.8} & \textbf{68.0} \\
DIVER-4B-1020     & 30.6             & \underline{72.8} & \underline{72.1} & \underline{61.9} & \underline{59.9} & \underline{56.7} & \underline{63.8} & \underline{58.8} & \underline{63.7} \\
DIVER-4B          & 28.9             & 67.3 & 71.2 & 53.4 & \underline{60.4} & 53.1 & 60.3 & 53.7 & 59.9 \\
RTriever-4B (ours) & 27.7             & 63.1 & 64.7 & 53.6 & 50.1 & 49.2 & 52.7 & 53.9 & 55.3 \\
INF-Retriever-Pro (7B) & 26.3             & 62.6 & 64.3 & 50.7 & 51.9 & 46.1 & 50.1 & 50.6 & 53.8 \\
Qwen3-8B          & 23.7             & 52.7 & 58.6 & 46.0 & 46.6 & 48.4 & 52.8 & 41.0 & 49.5 \\
\instructor-XL (1.5B) & 18.9             & 45.1 & 54.6 & 44.2 & 45.9 & 44.5 & 47.5 & 42.3 & 46.3 \\
OpenAI-Embed-3L   & 17.9             & 53.5 & 55.2 & 44.7 & 44.6 & 37.6 & 44.2 & 40.8 & 45.8 \\
GTE-7B            & 22.5             & 60.3 & 62.5 & 39.1 & 37.4 & 41.8 & 40.0 & 37.5 & 45.5 \\
GritLM (7B)       & 21.0             & 50.6 & 58.3 & 40.3 & 37.6 & 43.8 & 38.8 & 39.2 & 44.1 \\
ReasonIR-8B       & 24.4             & 42.6 & 52.2 & 34.9 & 35.6 & 40.9 & 45.7 & 35.1 & 41.0 \\
BM25              & 14.5             & 41.9 & 49.1 & 40.2 & 30.6 & 39.2 & 40.2 & 40.7 & 40.3 \\
EmbeddingGemma-300M & 18.9            & 48.7 & 55.8 & 32.6 & 32.7 & 35.3 & 35.2 & 35.7 & 39.4 \\
\bottomrule
\end{tabular}
}
\caption{Performance of retrievers under \ours static retrieval setting. We report $\mathrm{\alpha\text{-}nDCG}@25$ scores across all subsets, along with the NDCG@10 score for the \bright results. Models are ranked by their overall performance. The best and second-best models for each subset are shown in bold and underlined, respectively.}
\label{tab:model_performance}
\end{table*}

%% file: tables/fix-round.tex
\begin{table*}[!t]
\centering
\small
\resizebox{\linewidth}{!}{
\begin{tabular}{l*{9}{C{1.3cm}}}
\toprule
\multirow{2}{*}{\textbf{Model}} &
\multicolumn{3}{c}{\textbf{Round 1}} &
\multicolumn{3}{c}{\textbf{Round 2}} &
\multicolumn{3}{c}{\textbf{Round 3}} \\
\cmidrule(lr){2-4}\cmidrule(lr){5-7}\cmidrule(lr){8-10}
& \makecell{\textbf{$\bm{\alpha}$-nDCG}\\(0-100)} & \makecell{\textbf{Compl.}\\(1-5)} & \makecell{\textbf{Overall}\\(1-5)}
& \makecell{\textbf{$\bm{\alpha}$-nDCG}\\(0-100)} & \makecell{\textbf{Compl.}\\(1-5)} & \makecell{\textbf{Overall}\\(1-5)}
& \makecell{\textbf{$\bm{\alpha}$-nDCG}\\(0-100)} & \makecell{\textbf{Compl.}\\(1-5)} & \makecell{\textbf{Overall}\\(1-5)} \\
\midrule
BGE-Reasoner-8B       & \textbf{56.66}    & \textbf{4.14}    & \textbf{3.95}    & \textbf{61.35}    & \textbf{4.40}    & \textbf{4.26}    & \textbf{63.04}    & \textbf{4.42}    & \textbf{4.31} \\
DIVER-4B              & \underline{46.75} & 3.95             & 3.78             & \underline{50.29} & 4.21             & 4.12             & \underline{53.08} & \underline{4.38} & \underline{4.29} \\
RTriever-4B (ours)    & 44.22             & \underline{3.98} & \underline{3.85} & 48.17             & 4.17             & 4.12             & 50.79             & 4.37             & 4.25 \\
GTE-7B                & 45.88             & 3.93             & 3.84             & 49.80             & 4.17             & \underline{4.13} & 52.68             & 4.33             & 4.23 \\
DIVER-4B-1020         & 44.28             & 3.90             & 3.77             & 49.21             & 4.22             & 4.06             & 51.56             & 4.33             & 4.16 \\
\instructor-XL        & 35.24             & 3.54             & 3.50             & 40.15             & 4.09             & 3.97             & 43.48             & 4.26             & 4.14 \\
BM25                  & 44.46             & 3.82             & 3.67             & 48.05             & 4.19             & 4.05             & 51.48             & 4.25             & 4.12 \\
ReasonIR-8B           & 37.14             & 3.81             & 3.65             & 41.28             & 4.06             & 3.93             & 44.90             & 4.16             & 4.11 \\
Qwen3-8B              & 44.11             & 3.81             & 3.74             & 48.77             & \underline{4.25} & 4.11             & 51.75             & 4.26             & 4.10 \\
GritLM (7B)           & 40.28             & 3.89             & 3.74             & 43.99             & 4.19             & 4.03             & 47.05             & 4.24             & 4.07 \\
\bottomrule
\end{tabular}
}
\caption{Performance of retrievers under \ours fixed-round agentic evaluation setting (GPT-5-mini agent). Because each round returns top-5 passages, we report $\alpha$-nDCG at the cumulative cut-off: $k{=}5$ after round 1, $k{=}10$ after round 2, and $k{=}15$ after round 3. \textbf{Compl}: reasoning completeness of the final response; \textbf{Overall}: overall quality of the final response. Rows are ordered by round-3 Overall quality (descending).}
\label{tab:roundwise}
\end{table*}

%% file: tables/dynamic-round.tex
\begin{wraptable}{r}{0.48\linewidth}
\centering
\small
\setlength{\tabcolsep}{3pt}
\vspace{-1.2em}
\resizebox{\linewidth}{!}{
\begin{tabular}{l l C{1.0cm}*{3}{C{0.85cm}}}
\toprule
\textbf{Model} & \textbf{Agent} & \textbf{\#R}  & \textbf{Compl.} & \textbf{Overall} & \textbf{AER} \\
\midrule
\multirow{2}{*}{BGE-Reasoner-8B}     & GPT-5-mini & \textbf{5.10} & \textbf{4.63} & 4.43 & \textbf{3.65} \\
                                     & Qwen3.5    & \textbf{4.14} & \underline{4.22} & 3.99 & \textbf{3.44} \\
\midrule
\multirow{2}{*}{RTriever-4B (ours)}  & GPT-5-mini & 6.01 & 4.53 & 4.43 & 3.51 \\
                                     & Qwen3.5    & \underline{4.89} & \textbf{4.26} & \textbf{4.06} & \underline{3.38} \\
\midrule
\multirow{2}{*}{BM25}                & GPT-5-mini & 5.73 & 4.50 & 4.42 & 3.53 \\
                                     & Qwen3.5    & 5.21 & 4.19 & 4.01 & 3.31 \\
\midrule
\multirow{2}{*}{DIVER-4B}            & GPT-5-mini & 5.91 & 4.57 & 4.46 & 3.53 \\
                                     & Qwen3.5    & 5.56 & 4.20 & \underline{4.02} & 3.29 \\
\midrule
\multirow{2}{*}{DIVER-4B-1020}       & GPT-5-mini & \underline{5.61} & 4.54 & 4.43 & \underline{3.56} \\
                                     & Qwen3.5    & 6.57 & 4.18 & 3.96 & 3.11 \\
\midrule
\multirow{2}{*}{Qwen3-8B}            & GPT-5-mini & 6.27 & 4.52 & \underline{4.49} & 3.50 \\
                                     & Qwen3.5    & 6.16 & 4.15 & 3.93 & 3.14 \\
\midrule
\multirow{2}{*}{GTE-7B}              & GPT-5-mini & 6.67 & \underline{4.62} & \textbf{4.51} & 3.44 \\
                                     & Qwen3.5    & 5.83 & 4.13 & 3.95 & 3.18 \\
\midrule
\multirow{2}{*}{GritLM (7B)}         & GPT-5-mini & 7.03 & 4.51 & 4.47 & 3.36 \\
                                     & Qwen3.5    & 6.48 & 4.08 & 3.85 & 3.08 \\
\midrule
\multirow{2}{*}{ReasonIR-8B}         & GPT-5-mini & 7.17 & 4.48 & 4.42 & 3.31 \\
                                     & Qwen3.5    & 6.99 & 4.10 & 3.93 & 3.08 \\
\midrule
\multirow{2}{*}{\instructor-XL} & GPT-5-mini & 7.14 & 4.46 & 4.35 & 3.26 \\
                                     & Qwen3.5    & 6.93 & 4.05 & 3.88 & 3.04 \\
\bottomrule
\end{tabular}
}
\caption{Performance of retrievers under \ours adaptive-round agentic evaluation setting, with GPT-5-mini and Qwen3.5-122B-A10B as backbones. Rows are ordered by the mean AER across the two agents.}
\label{tab:iter_results}
\end{wraptable}

%% file: main/6-conclusion.tex
\section{Conclusion}
We introduce \ours, a reasoning-intensive retrieval benchmark with multi-aspect gold sets and agentic retriever-in-the-loop protocols, together with \model-Synth, an aspect-decomposed synthetic corpus for training retrievers.
Our experiments show that aspect-aware evaluation exposes retrieval behaviors hidden by single-passage metrics, and that static retrieval quality does not always predict utility inside agentic search loops.
On the training side, \model-4B substantially improves over its Qwen3-Embedding-4B base, suggesting that complementary positives and positive-conditioned negatives provide useful supervision for reasoning-intensive retrieval. These findings call for retrievers built and evaluated around full evidence portfolios rather than individual passages.

%% file: main/limitations.tex
\section*{Limitations and Future Work}
\ours builds upon the StackExchange subset of \bright, which currently covers seven expert domains. However, this scope may not fully capture the diversity and complexity of real-world, reasoning-intensive retrieval scenarios. Future research could extend our work by incorporating a broader range of expert domains~\cite{zhao-etal-2024-findver, shen-etal-2026-patient} to enhance representativeness.
Additionally, the number of examples in \ours remains limited due to the high cost of human expert annotation. Future work could explore semi-automated or hybrid annotation pipelines to scale up data creation while maintaining quality and reliability.
Finally, our use of \model-Synth is intentionally simple: we fine-tune \model-4B on a subset of one-positive-one-negative triplets. The full training corpus enables further exploration of multi-positive objectives~\cite{Wang2026TrainingDR,Esfandiarpoor2025BeyondCL}, aspect-aware sampling~\cite{Dou2025FLeWFA}, negative curricula~\cite{Zhou2025ARKAR,Moreira2025ImprovingTE}, and how synthetic evidence structure affects reasoning-intensive retrievers~\cite{Feng2026TheWO}.

%% file: appendix/main.tex
\clearpage
\input{appendix/aspect_weight_rubric}

\input{appendix/data_quality_control}

\input{appendix/static_metrics}

\input{appendix/reference_answer_validation}

\section{System Prompts Template}\label{app:prompt}
Figures~\ref{fig:deep-research-prompt}, \ref{fig:final-response-prompt}, \ref{fig:reference-answer-prompt}, and \ref{fig:judge-prompt} illustrate the prompt templates used in our experiments (agent main prompt, fixed-round response generation, reference-answer generation, and LLM-as-Judge scoring, respectively).
\input{figures_tex/prompt}

\clearpage
\section{Agent Decoding and Search Configuration}\label{app:agent-config}
All agentic experiments share the LLM-side decoding and tool-side search settings listed below across both agent backends, so that performance differences across retrievers are attributable to retrieval quality rather than to generation variability or agent stopping behavior.

\paragraph{Agent Backends.} We evaluate every retriever with two complementary backends, both queried through an OpenAI Responses-API-compatible interface:
\begin{itemize}[leftmargin=1.5em,itemsep=0pt]
    \item \textbf{GPT-5-mini-08-07}, accessed through the Azure OpenAI Responses API. We set \texttt{reasoning\_effort}=\texttt{medium} and leave \texttt{temperature} and \texttt{top\_p} at the model's defaults (not overridden). We allow up to 30{,}000 output tokens per search/reasoning turn and 10{,}000 tokens for the final answer.
    \item \textbf{Qwen3.5-122B-A10B-GPTQ-Int4}~\cite{yang2025qwen3technicalreport}, self-hosted via vLLM~v0.19.1 with \texttt{--reasoning-parser qwen3} and \texttt{--tool-call-parser qwen3\_coder}, exposing a stateless \texttt{/v1/responses} endpoint. We allow up to 25{,}600 output tokens per turn and 12{,}800 tokens for the final answer; \texttt{temperature} and \texttt{top\_p} are left at the served defaults.
\end{itemize}
Both backends use identical system/user prompts, tool schema, top-$k$ setting, snippet-truncation budget, and round/iteration limits.

\paragraph{Search Tool.} Each call to the search tool returns the top-$5$ retrieved documents. The text of each returned document is truncated to 2{,}048 tokens with the \texttt{Qwen/Qwen3-0.6B} tokenizer before being shown to the agent, so that a single long document cannot dominate the context window. The agent is given a single \texttt{search} tool (no separate \texttt{get\_document} tool); the tool schema and prompt template are shared across backends and across retrievers.

\paragraph{Interaction Budgets.} Under the \emph{fixed-round} protocol the agent runs the conversation through search rounds $r\in\{1,2,3\}$ and, at each of those rounds, generates a final answer conditioned on all documents retrieved up to round $r$ (a safety cap of $10$ tool-calling iterations is set but is never the binding constraint, since termination is triggered at round~$3$). Under the \emph{adaptive-round} protocol the agent decides when to stop on its own, up to a hard limit of $100$ rounds.

\paragraph{Sampling.} For both protocols we evaluate on a fixed sample of $175$ queries (25 per task, drawn with a fixed random seed from the 739-query \ours pool); every retriever is evaluated on the identical sample, so cross-retriever differences are not confounded by question selection.

\clearpage
\onecolumn
\section{Data Examples}\label{app:data-example}
\input{appendix/bright_pro_example}

\clearpage
\section{Experiment Results}\label{app:exp}
\input{tables/appendix_metric}

\input{tables/fix-round-qwen35}

\input{appendix/case_study}

%% file: appendix/aspect_weight_rubric.tex
\section{Aspect Importance Rubric}
\label{app:rebric}

To reflect unequal contributions across reasoning aspects, annotators assign each aspect a Likert-scale score from 1 to 5 based on its importance to producing a correct and defensible final answer. The scores are then normalized into weights $w_a \in [0,1]$ with $\sum_a w_a = 1$. Table~\ref{tab:importance_rubric} details the grading rubric used during annotation.

\begin{table*}[h]
\centering
\small
\setlength{\tabcolsep}{10pt}
\begin{tabular}{c|p{0.82\linewidth}}
\toprule
\textbf{Score} & \textbf{Interpretation} \\
\midrule
\textbf{5} & \textbf{Dominant:} This aspect captures the core reasoning required for a correct answer. It directly determines whether the response’s main claim or logic is valid. \\[4pt]
\textbf{4} & \textbf{Strong:} The aspect provides key supporting reasoning that significantly improves correctness or interpretive depth, though not the single decisive factor. \\[4pt]
\textbf{3} & \textbf{Moderate:} The aspect is necessary to develop a coherent or complete explanation, but its omission would only moderately weaken correctness. \\[4pt]
\textbf{2} & \textbf{Supporting:} The aspect refines or contextualizes the reasoning, helping link ideas or strengthen justification but is secondary in determining correctness. \\[4pt]
\textbf{1} & \textbf{Peripheral:} The aspect has minimal influence compared with others but still contributes meaningfully to understanding or completeness. \\
\bottomrule
\end{tabular}
\caption{Rubric for assigning weights (1–5) among meaningful reasoning aspects.}
\label{tab:importance_rubric}
\end{table*}

%% file: appendix/data_quality_control.tex
\section{Data Quality Control Protocol}\label{app:data-quality}
This section expands on the second-annotator review summarized in the main text.

\paragraph{Aspect Review.}
For each example, reviewers carefully assess the reasoning aspects to verify that they are comprehensive, balanced, and conceptually sound. The aspects should neither be overly granular nor too general, and should collectively capture the complete reasoning process behind the question. Reviewers also verify that the assigned weight accurately reflects each aspect's contribution to the reasoning, adjusting weights when initial annotations overemphasize minor details or undervalue critical components.

\paragraph{Positive Document Review.}
After verifying the reasoning aspects, the reviewers re-examine each positive document to confirm that it provides explicit, credible, and directly relevant evidence for its assigned aspect. Reviewers remove redundant or loosely related content and ensure that every passage offers concrete reasoning support rather than surface-level associations.

%% file: appendix/static_metrics.tex
\section{Static Retrieval Metrics: Notation and Formulas}\label{app:static-metrics}

We provide the formal definitions of the four metrics used in the static retrieval evaluation.
As discussed in Section~\ref{sec:data-aspect}, each query $q$ is annotated with $m$ reasoning aspects $\{a_1,\dots,a_m\}$, with each aspect assigned a nonnegative weight $w_j$ representing its importance to the overall reasoning process. Let $\mathcal{G}$ be the set of gold passages and let $a:\mathcal{G}\to\{1,\dots,m\}$ map each gold passage to its unique aspect. Given a ranked list $\langle d_1,\dots,d_k\rangle$, $r\in\{1,\dots,k\}$ denotes the rank position and $d_r$ is the document at rank $r$. We define $C_j(k)=\sum_{r=1}^{k}\mathrm{rel}_{rj}$ as the count of retrieved gold items for aspect $j$ within the top $k$, and the binary relevance $\mathrm{rel}_r=\ind{d_r\in\mathcal{G}}$, so that the aspect-specific relevance becomes
\begin{equation}
\begin{split}
\mathrm{rel}_{rj} &= \ind{d_r\in\mathcal{G} \land a(d_r)=j} \\
                  &= \mathrm{rel}_r\,\ind{a(d_r)=j}.
\end{split}
\end{equation}

\paragraph{\andcgb.} \andcg~\cite{andcg} explicitly models aspect-level diversity, rewarding retrievers that cover multiple reasoning aspects rather than retrieving redundant passages from the same subtopic. The gain at rank $r$ is
\begin{equation}
G(r)=\sum_{j=1}^{m} w_j \,\mathrm{rel}_{rj}\,(1-\alpha)^{C_j(r-1)},
\end{equation}
where $\alpha\in[0,1]$ controls the novelty penalty (we set $\alpha=0.5$ throughout the paper). Accumulating with a logarithmic discount,
\begin{equation}
\mathrm{DCG}_{\alpha}@k=\sum_{r=1}^{k}\frac{G(r)}{\log_2(r+1)}.
\end{equation}
For normalization, we compute $\mathrm{IDCG}_{\alpha}@k$ on the gold pool using the same gain definition and a greedy maximization over $k$ positions, yielding
\begin{equation}
\mathrm{\alpha\text{-}nDCG}@k=\frac{\mathrm{DCG}_{\alpha}@k}{\mathrm{IDCG}_{\alpha}@k}.
\end{equation}

\paragraph{Weighted Aspect Recall.} To directly capture aspect coverage, we report a weighted aspect recall that credits each aspect once it has been covered at least once:
\begin{equation}
\mathrm{A\text{-}Recall}@k \;=\; \sum_{j=1}^{m} w_j \cdot \ind{C_j(k)\ge 1}.
\end{equation}

\paragraph{Recall@$k$ and NDCG@$k$.} As complementary metrics that ignore aspect structure, we additionally report the standard Recall@$k$ (fraction of gold passages within the top $k$) and NDCG@$k$ (with binary relevance $\mathrm{rel}_r$).

%% file: appendix/reference_answer_validation.tex
\section{Reference Answer Validation}\label{app:ref-answer-validation}
We use GPT-5 with a high reasoning effort setting to generate one citation-grounded reference answer per query. The model is given the human-annotated reasoning aspects together with the full content of the positive passages linked to each aspect, and is prompted to produce a comprehensive answer that cites the supporting passages.

To verify that the synthesized reference answers are reliable enough to serve as the gold target for LLM-as-Judge scoring, we randomly sample $40$ examples and have the same annotators who originally worked on each example compose a citation-grounded answer from scratch. One of the authors then blindly rates the model- and human-written answers on a five-point Likert scale across answer correctness and completeness. The model-generated answers achieve an average completeness score of $4.80$ and an overall quality score of $4.85$, closely matching the human-written answers ($4.75$ for completeness and $4.90$ for overall quality). The match indicates that the synthesized references are of sufficiently high quality to serve as reliable targets for LLM-as-Judge scoring.

%% file: figures_tex/prompt.tex
\begin{figure*}[h]
\begin{tcolorbox}[colback=black!7.5!white, colframe=black!80!white, title=Deep Research Main Prompt, fontupper=\footnotesize, fonttitle=\footnotesize]

Question: \{Question\}
\\
You are a research agent. Your task is to answer the question by actively using the provided Search Tool.

Use the search tool iteratively for many turns. But in each turn, you should only use the search tool once.

Refine your queries based on previous results to broaden coverage and fill knowledge gaps.

Stop searching only once you have gathered a comprehensive and multi-perspective set of evidence.

Your final response must integrate information from different angles, supported by multiple sources. You must base your answer solely on the retrieved evidence documents—do not use any prior or external knowledge.\\

Your final response should be in the following format:\\
Answer: \{\{Your final answer. You should cite your evidence documents inline by enclosing their docids in square brackets at the end of sentences. For example, [20].\}\}\\
Confidence: \{\{Your confidence score between 0\% and 100\% for your answer\}\}\\
\end{tcolorbox}

\caption{Prompt to run deep research agent.}
\label{fig:deep-research-prompt}
\end{figure*}

\begin{figure*}[h]
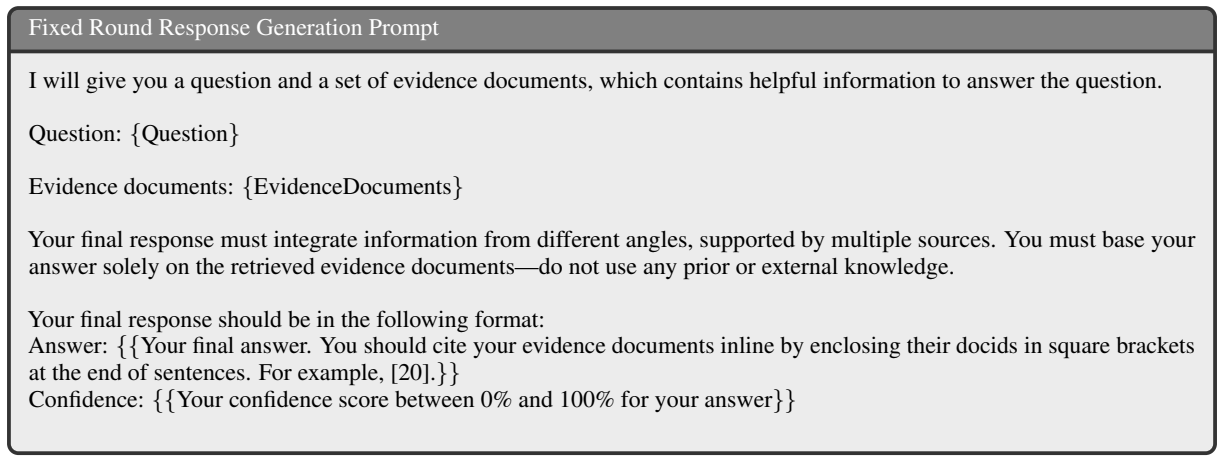

\begin{tcolorbox}[colback=black!7.5!white, colframe=black!80!white, title=Fixed Round Response Generation Prompt, fontupper=\footnotesize, fonttitle=\footnotesize]

I will give you a question and a set of evidence documents, which contains helpful information to answer the question.\\

Question: \{Question\}\\

Evidence documents: \{EvidenceDocuments\}\\

Your final response must integrate information from different angles, supported by multiple sources. You must base your answer solely on the retrieved evidence documents—do not use any prior or external knowledge.\\

Your final response should be in the following format:\\
Answer: \{\{Your final answer. You should cite your evidence documents inline by enclosing their docids in square brackets at the end of sentences. For example, [20].\}\}\\
Confidence: \{\{Your confidence score between 0\% and 100\% for your answer\}\}\\
\end{tcolorbox}

\caption{Prompt to generate the final response after a fixed round of retrieval. At each fixed round $r\!\in\!\{1,2,3\}$, \texttt{\{EvidenceDocuments\}} is the concatenation of all documents retrieved through round $r$.}
\label{fig:final-response-prompt}
\end{figure*}

\begin{figure*}[h]
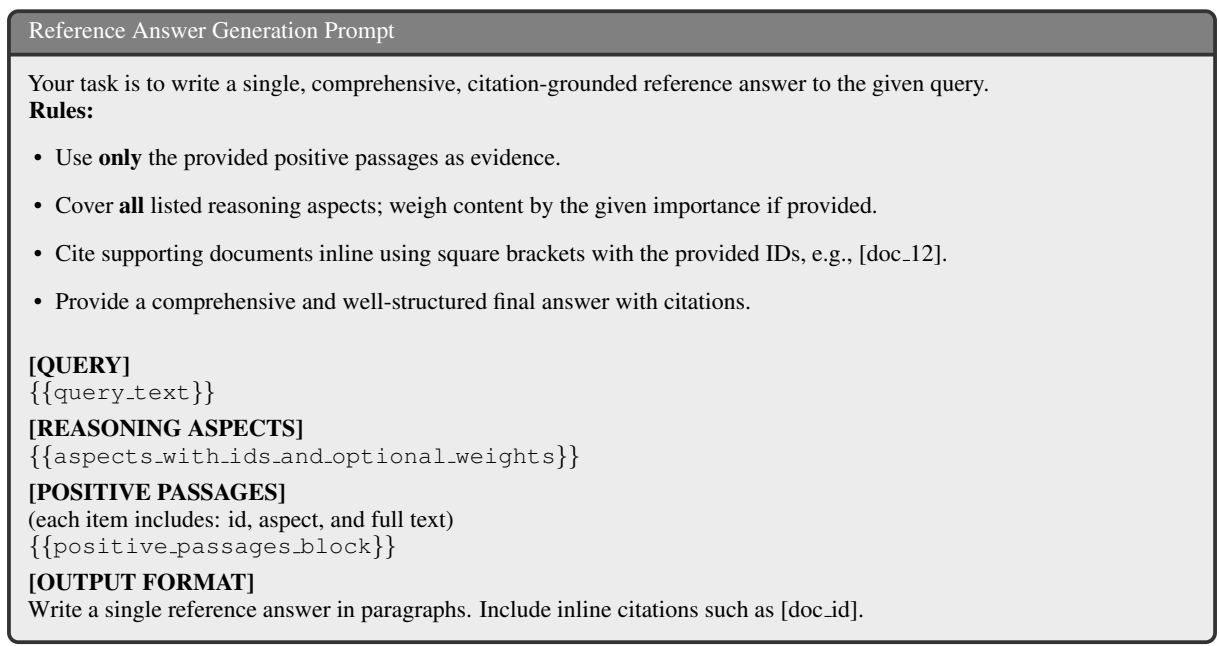

\begin{tcolorbox}[colback=black!7.5!white, colframe=black!80!white, title=Reference Answer Generation Prompt, fontupper=\footnotesize, fonttitle=\footnotesize]

Your task is to write a single, comprehensive, citation-grounded reference answer to the given query.

\textbf{Rules:}
\begin{itemize}[leftmargin=10pt]
    \item Use \textbf{only} the provided positive passages as evidence.
    \item Cover \textbf{all} listed reasoning aspects; weigh content by the given importance if provided.
    \item Cite supporting documents inline using square brackets with the provided IDs, e.g., [doc\_12].
    \item Provide a comprehensive and well-structured final answer with citations.
\end{itemize}

\vspace{6pt}
\textbf{[QUERY]}\\
\texttt{\{\{query\_text\}\}}

\vspace{4pt}
\textbf{[REASONING ASPECTS]}\\
\texttt{\{\{aspects\_with\_ids\_and\_optional\_weights\}\}}

\vspace{4pt}
\textbf{[POSITIVE PASSAGES]}\\
(each item includes: id, aspect, and full text)\\
\texttt{\{\{positive\_passages\_block\}\}}

\vspace{4pt}
\textbf{[OUTPUT FORMAT]}\\
Write a single reference answer in paragraphs. Include inline citations such as [doc\_id].

\end{tcolorbox}

\caption{Prompt for reference answer generation, showing input structure and output specification.}
\label{fig:reference-answer-prompt}
\end{figure*}

\begin{figure*}[h]
\begin{tcolorbox}[colback=black!7.5!white, colframe=black!80!white, title=LLM-as-Judge Scoring Prompt (System), fontupper=\footnotesize, fonttitle=\footnotesize]

You are an expert evaluator grading a research-assistant's answer.\\

For each example you receive:
\begin{itemize}[leftmargin=12pt,itemsep=0pt]
    \item \textbf{QUESTION} --- a query from a specialized StackExchange community.
    \item \textbf{REASONING\_ASPECTS} --- a list of key sub-questions / premises a correct answer must address. Each aspect has a short id (\texttt{a1}, \texttt{a2}, \dots). Treat the list as the authoritative rubric.
    \item \textbf{REFERENCE\_ANSWER} --- a citation-grounded answer produced from the gold evidence passages. This is the high-quality target.
    \item \textbf{SYSTEM\_ANSWER} --- the answer produced by the system under evaluation.
\end{itemize}

\textbf{Step 1 --- score EACH aspect on a 3-point coverage scale:}
\begin{itemize}[leftmargin=24pt,itemsep=0pt]
    \item[1.0] fully addressed with specific, correctly-supported claims
    \item[0.5] partially addressed (mentioned but shallow, OR addressed with notable inaccuracies, OR right idea but missing critical detail)
    \item[0.0] not addressed, off-topic, or factually wrong
\end{itemize}
You MUST grade every aspect id given in REASONING\_ASPECTS --- do not skip any.

\vspace{4pt}
\textbf{Step 2 --- assign one holistic \texttt{overall\_quality} score (1--5 integer)} for SYSTEM\_ANSWER relative to REFERENCE\_ANSWER (correctness, structure, citations, coherence, no hallucinations):
\begin{itemize}[leftmargin=24pt,itemsep=0pt]
    \item[5] matches or exceeds REFERENCE\_ANSWER
    \item[4] slightly worse but still correct and well-structured
    \item[3] correct but less thorough or less clearly structured
    \item[2] partially correct with notable issues
    \item[1] mostly wrong or hallucinated
\end{itemize}

\vspace{2pt}
Return STRICTLY a single JSON object, no markdown fences, no prose outside the object:\\[2pt]
\texttt{\{"aspect\_scores": \{"a1": <0$|$0.5$|$1>, "a2": <0$|$0.5$|$1>, ...\}, "overall\_quality": <int 1-5>, "justification": "<1-2 sentence rationale>"\}}

\tcblower
\textbf{User message}\\[2pt]
\texttt{QUESTION:}\\
\texttt{\{question\}}

\vspace{2pt}
\texttt{REASONING\_ASPECTS:}\\
\texttt{\{aspects\_block\}}\hfill\textit{(bulleted: \texttt{- a1: <text>, w=<likert>})}

\vspace{2pt}
\texttt{REFERENCE\_ANSWER:}\\
\texttt{\{reference\_answer\}}

\vspace{2pt}
\texttt{SYSTEM\_ANSWER:}\\
\texttt{\{system\_answer\}}
\end{tcolorbox}

\caption{Prompt for LLM-as-Judge scoring of system responses. The judge returns one $\{0, 0.5, 1\}$ coverage score per aspect (\texttt{aspect\_scores}) plus a 1--5 holistic \texttt{overall\_quality} score; the 1--5 \texttt{reasoning\_completeness} reported in the main text is computed post-hoc as $\mathrm{round}(4 \cdot \overline{w} + 1)$, where $\overline{w}$ is the weight-normalized mean of the per-aspect coverage scores using the Likert aspect weights.}
\label{fig:judge-prompt}
\end{figure*}

%% file: appendix/bright_pro_example.tex
\begin{table*}[h]
\centering
\small
\begin{tabularx}{\linewidth}{@{}X@{}}
\toprule
\textbf{Query}:
Did neanderthals need vitamin C from the diet? Modern humans need to get vitamin C from the diet, because we do not make it ourselves. Did neanderthals produce vitamin C? At what point of evolution exactly was this ability lost?\\
\midrule

\textbf{Reasoning Aspect 1 (weight = 0.27)}\\
Vitamin C biosynthesis in animals requires the enzyme L\nobreakdash-gulonolactone oxidase (encoded by the \textit{GULO}/\textit{GLO} gene), which catalyzes the final step of the pathway.\\[4pt]
\textbf{Positive Documents}\\
\textbf{Doc 1:} (...abbrev...) Review of vertebrate vitamin C synthesis losses in teleost fishes, anthropoid primates, guinea pigs, and some bats and Passeriformes; in every studied case the inability to synthesize is caused by mutations in the \textit{GLO} gene that codes for the enzyme catalyzing the last step of biosynthesis (the figure of the biosynthesis pathway labels enzyme 6 as L\nobreakdash-gulonolactone oxidase, GLO, EC 1.1.3.8). (...abbrev...)\\
\textbf{Doc 2:} (...abbrev...) ``Why are mutations limited to \textit{GLO} genes?'' --- because losing \textit{GLO} only affects vitamin C production, whereas losing genes for other enzymes in the same pathway (e.g.\ gluconolactonase, EC 3.1.1.17) would also disrupt caprolactam degradation and the pentose phosphate pathway; the \textit{GLO} gene is therefore ``predisposed'' to being lost since it makes a single compound unnecessary for other pathways. (...abbrev...)\\

\midrule

\textbf{Reasoning Aspect 2 (weight = 0.27)}\\
Anthropoid primates carry a mutated, nonfunctional \textit{GULOP} pseudogene, preventing endogenous vitamin C synthesis and forcing dietary intake.\\[4pt]
\textbf{Positive Documents}\\
\textbf{Doc 1:} (...abbrev...) The nonfunctional \textit{GULOP} pseudogene maps to human chromosome 8p21; loss of \textit{GULO} activity occurred independently in some bats, in guinea pigs, and in the haplorrhine suborder of primates (which includes humans). The mutated remnant is still present in guinea pig and human genomes. (...abbrev...)\\

\midrule

\textbf{Reasoning Aspect 3 (weight = 0.27)}\\
Loss of functional \textit{GULO} in primates occurred about 61--63 million years ago, around the haplorrhine--strepsirrhine split.\\[4pt]
\textbf{Positive Documents}\\
\textbf{Doc 1:} (...abbrev...) Anthropoid primates have lost 7 of the 12 exons of \textit{GLO}; inactivation dates derived from comparing functional and nonfunctional sequences place the event at $\sim$61\,MYA in anthropoid primates and $\sim$14\,MYA in guinea pigs, consistent with phylogenetic estimates. (...abbrev...)\\

\midrule

\textbf{Reasoning Aspect 4 (weight = 0.18)}\\
Neanderthals belong to the haplorrhine hominid lineage and share the degenerated \textit{GULOP} sequence with modern humans, so they too could not synthesize vitamin C.\\[4pt]
\textbf{Positive Documents}\\
\textbf{Doc 1:} (...abbrev...) Haplorrhini and strepsirrhini primates diverged $\sim$70\,MYA; \textit{GULOP} is found in all haplorrhini, while strepsirrhini retain a functional \textit{GULO}. Neanderthal genomic data place them within the haplorrhine hominids alongside \textit{Pongo}, \textit{Gorilla}, \textit{Pan}, and \textit{Homo}. (...abbrev...)\\

\bottomrule
\end{tabularx}
\caption{\ours~Biology Example.}
\label{tab:biology-example}
\end{table*}

\begin{table*}[!t]
\centering
\small
\begin{tabularx}{\linewidth}{@{}X@{}}
\toprule
\textbf{Query}:\\
I have never understood why earth's inner core is solid. Considering that the inner core is made of an iron--nickel alloy (melting point around $1350\,^\circ$C to $1600\,^\circ$C) and the temperature of the inner core is approximately $5430\,^\circ$C (about the temperature of the surface of the sun). Since Earth's core is nearly 3--4 times the melting point of iron--nickel alloys, how can it possibly be solid?\\
\midrule
\textbf{Reasoning Aspect 1 (weight = 0.30)}\\
For most materials the melting point rises with pressure: the solid phase is denser than the liquid, so higher temperatures are required to overcome intermolecular forces and melt the material at high pressure.\\[4pt]
\textbf{Positive Documents}\\
\textbf{Doc 1:} (...abbreviation...) Pressure--temperature phase diagrams show that the melting curve generally has a positive slope for most substances (water being the anomalous exception with a negative slope), so increasing pressure raises the melting temperature. (...abbreviation...)\\
\midrule
\textbf{Reasoning Aspect 2 (weight = 0.30)}\\
Earth's inner core experiences extreme pressures of $\sim$330--360\,GPa, which greatly elevate the melting temperature of iron and iron--nickel alloys above the local geotherm.\\[4pt]
\textbf{Positive Documents}\\
\textbf{Doc 1:} (...abbreviation...) Wikipedia article on Earth's inner core: pressure ranges from $\sim$330 to 360\,GPa (3.3--3.6 million atm); ``iron can be solid at such high temperatures only because its melting temperature increases dramatically at pressures of that magnitude (see the Clausius--Clapeyron relation)''; the inner core is solid in accordance with the Simon--Glatzel equation. (...abbreviation...)\\
\textbf{Doc 2:} (...abbreviation...) Description of Earth's outer and inner core: ``The only reason why iron and other heavy metals can be solid at such high temperatures is because their melting temperatures dramatically increase at the pressures present there, which ranges from about 330 to 360 gigapascals''; the outer core remains liquid because it ``is not under enough pressure to be solid'' despite a similar iron--nickel composition. (...abbreviation...)\\
\midrule
\textbf{Reasoning Aspect 3 (weight = 0.20)}\\
Experimental and theoretical studies estimate iron's melting temperature near inner-core boundary pressures to be roughly 5,500--7,000\,K, well above the local temperature.\\[4pt]
\textbf{Positive Documents}\\
\textbf{Doc 1:} (...abbreviation...) Review article: ``Previous studies give a range of iron melting temperatures from 5500 to 7000\,K at the ICB pressure'' across extrapolations from static laser-heated diamond-anvil-cell measurements (Anzellini et al., 2013; Morard et al., 2018), dynamic shock-wave compressions (Brown \& McQueen, 1986; Yoo et al., 1993), and ab initio theoretical calculations (Alf\`{e}, 2009; Bouchet et al., 2013). (...abbreviation...)\\
\textbf{Doc 2:} (...abbreviation...) Companion treatment of the iron melting curve at ICB ($\sim$330\,GPa): same 5500--7000\,K range from static, dynamic, and theoretical studies; specific shock-wave determinations include 5600(500)\,K at $\sim$243\,GPa (Brown \& McQueen) and 6830(500)\,K and 7800(500)\,K at the ICB (Yoo et al.; Bass et al.). (...abbreviation...)\\
\midrule
\textbf{Reasoning Aspect 4 (weight = 0.20)}\\
The inner core is solid while the outer core is liquid because the pressure-driven melting curve crosses the geotherm at the inner-core boundary.\\[4pt]
\textbf{Positive Documents}\\
\textbf{Doc 1:} (...abbreviation...) Inner core: $\sim$1220\,km radius, iron--nickel alloy, surface temperature $\sim$5430$\,^\circ$C ($\sim$ Sun's photosphere); the high-pressure solid phase is inferred from seismic waves and Earth's magnetic field. (...abbreviation...)\\
\textbf{Doc 2:} (...abbreviation...) ``The isentropic temperature profile in the liquid iron alloy in the outer core of Earth intersects the iron melting curve at $\sim$330\,GPa, forming the outer boundary of the solid inner core'' --- in other words, the geotherm crosses iron's pressure-elevated melting curve at the ICB, leaving the deeper part solid and the shallower part liquid (where convection drives the geodynamo). (...abbreviation...)\\
\bottomrule
\end{tabularx}
\caption{\ours~Earth Science Example.}
\label{tab:earth-science-example}
\end{table*}

\begin{table*}[!t]
\centering
\small
\begin{tabularx}{\linewidth}{@{}X@{}}
\toprule
\textbf{Query}:\\
Is there any example for ``pluralistic ignorance'' in economics? I read in Bursztyn (2021) that misperceptions about others can be explained by stereotyping, motivated reasoning, and pluralistic ignorance --- the situation in which almost all members of a group privately reject group norms, yet believe that virtually all other members accept them. Could you give an example in economics or finance to make this concrete?\\
\midrule
\textbf{Reasoning Aspect 1 (weight = 0.25)}\\
Pluralistic ignorance arises when individuals privately hold preferences differing from perceived majority norms, leading them to publicly conform despite personal disagreement (``preference falsification'').\\[4pt]
\textbf{Positive Documents}\\
\textbf{Doc 1:} (...abbreviation...) Following Kuran (1995), ``public preference falsification'' refers to settings where people publicly express positions opposite to their privately held views; it differs from white lies (it brings disutility) and from self-deception (it is a social phenomenon driven by conformity to perceived majority preferences). (...abbreviation...)\\
\midrule
\textbf{Reasoning Aspect 2 (weight = 0.25)}\\
In Saudi Arabia, most young married men privately support women working outside the home but underestimate other men's support, suppressing their wives' labor-force participation.\\[4pt]
\textbf{Positive Documents}\\
\textbf{Doc 1:} (...abbreviation...) Bursztyn, Gonz\'{a}lez, and Yanagizawa-Drott (2020) document low female labor-force participation in Saudi Arabia ($\sim$15\%) alongside even lower observable employment of wives ($\sim$4\% in their sample); using a survey experiment they show that men privately support women working outside the home but systematically underestimate other men's support, a textbook case of pluralistic ignorance. (...abbreviation...)\\
\midrule
\textbf{Reasoning Aspect 3 (weight = 0.17)}\\
Tax-compliance research shows many taxpayers personally value honesty yet believe that others tolerate evasion, sustaining a mistaken descriptive norm of noncompliance.\\[4pt]
\textbf{Positive Documents}\\
\textbf{Doc 1:} (...abbreviation...) Wenzel (2001) and others find that self-reported compliance correlates with beliefs about others' compliance, and that most people believe one should be honest yet think other taxpayers are not honest --- a divergence between injunctive and descriptive norms that fits the pluralistic-ignorance pattern. (...abbreviation...)\\
\midrule
\textbf{Reasoning Aspect 4 (weight = 0.17)}\\
In studies of bribery, individuals routinely overestimate community approval of corrupt practices even though only small minorities actually endorse them.\\[4pt]
\textbf{Positive Documents}\\
\textbf{Doc 1:} (...abbreviation...) In a multi-state Nigerian study of pass-mark bribery, respondents in all states except Sokoto systematically overestimated the share of community members who hold positive normative beliefs about bribery: e.g., respondents in Adamawa and Enugu thought $\sim$40\% of their community held such beliefs, when in fact only 8.6\% and 11\% did; the same gap holds for traffic-bribery solicitation (perceived 40--50\% vs.\ actual 8--8.2\%). (...abbreviation...)\\
\midrule
\textbf{Reasoning Aspect 5 (weight = 0.17)}\\
On U.S.\ climate policy, supporters outnumber opponents two-to-one, yet 80--90\% of Americans underestimate national support for major mitigation policies, discouraging advocacy and collective action.\\[4pt]
\textbf{Positive Documents}\\
\textbf{Doc 1:} (...abbreviation...) Representative U.S.\ survey ($N\!=\!6{,}119$) finds a near-universal misperception (``false social reality''): 66--80\% of Americans support major climate mitigation policies, but average estimates of public support are only 37--43\%; the misperception holds across every state and demographic, illustrating pluralistic ignorance as a barrier to collective action on climate change. (...abbreviation...)\\
\bottomrule
\end{tabularx}
\caption{\ours~Economics Example.}
\label{tab:economics-example}
\end{table*}

\begin{table*}[!t]
\centering
\small
\begin{tabularx}{\linewidth}{@{}X@{}}
\toprule
\textbf{Query}:
What is the scientific term for the tendency to see familiar patterns in things that are actually something completely different? One of the most common examples is perceiving human faces in clouds, cars, and all kinds of objects. I'm looking for a low-level misperception that people are usually aware of on a higher cognitive level (i.e., they know they are not actually seeing a face).\\
\midrule

\textbf{Reasoning Aspect 1 (weight = 0.33)}\\
\emph{Pareidolia} is the perceptual tendency to impose meaningful interpretations on ambiguous (typically visual) stimuli, detecting objects or patterns where none actually exist.\\[4pt]
\textbf{Positive Documents}\\
\textbf{Doc 1:} (...abbreviation...) Encyclopedic definition: ``Pareidolia is the tendency for perception to impose a meaningful interpretation on a nebulous stimulus, usually visual, so that one detects an object, pattern, or meaning where there is none. Pareidolia is a type of apophenia.'' (...abbreviation...)\\
\textbf{Doc 2:} (...abbreviation...) Same definition expanded with examples and scope: ``a specific but common type of apophenia''; common examples include perceived images of animals, faces, or objects in cloud formations, faces in inanimate objects, and lunar pareidolia (Man in the Moon, Moon rabbit); the concept extends to hearing voices or music in random noise (e.g., from air conditioners or fans), and face pareidolia has been demonstrated in rhesus macaques. (...abbreviation...)\\
\midrule

\textbf{Reasoning Aspect 2 (weight = 0.22)}\\
Pareidolia is a specific subtype of \emph{apophenia} --- the broader human tendency to perceive meaningful connections within unrelated data.\\[4pt]
\textbf{Positive Documents}\\
\textbf{Doc 1:} (...abbreviation...) Apophenia (Klaus Conrad, 1958) is defined as the ``unmotivated seeing of connections accompanied by a specific feeling of abnormal meaningfulness''; introduced in the context of early-stage schizophrenia and distinguished from hallucination. (...abbreviation...)\\
\midrule

\textbf{Reasoning Aspect 3 (weight = 0.22)}\\
Neuroimaging shows face-like (pareidolic) stimuli activate face-processing regions such as the fusiform face area (FFA), similarly to real faces.\\[4pt]
\textbf{Positive Documents}\\
\textbf{Doc 1:} (...abbreviation...) A 2009 magnetoencephalography study showed that objects perceived as faces evoke an early ($\sim$165\,ms) FFA activation similar in time and location to that evoked by real faces ($\sim$130\,ms), supporting face perception of face-like objects as an early process rather than a late cognitive reinterpretation. (...abbreviation...)\\
\textbf{Doc 2:} (...abbreviation...) Kanwisher et al.\ identified the Fusiform Face Area (FFA) on the lateral mid-fusiform gyrus, with greater fMRI activation to faces than to letter strings or textures; Liu et al.\ further showed the FFA plays a crucial role in both real-face perception and face-pareidolia processing, with higher FFA activation during face pareidolia than during letter pareidolia, supporting an interaction between bottom-up (occipitotemporal) and top-down (frontal) processing. (...abbreviation...)\\
\midrule

\textbf{Reasoning Aspect 4 (weight = 0.22)}\\
Brain responses to illusory faces occur rapidly and may initially represent objects as faces before higher-level processing resolves the error.\\[4pt]
\textbf{Positive Documents}\\
\textbf{Doc 1:} (...abbreviation...) MEG plus fMRI multivariate decoding shows that illusory faces are first represented as more face-like than matched objects in face-selective occipito-temporal cortex, and only $\sim$100\,ms later are represented as more object-like --- consistent with a rapidly-deployed, broadly-tuned face-detection system whose output is subsequently corrected. (...abbreviation...)\\
\bottomrule
\end{tabularx}
\caption{\ours~Psychology Example.}
\label{tab:psychology-example}
\end{table*}

\begin{table*}[!t]
\centering
\small
\begin{tabularx}{\linewidth}{@{}X@{}}
\toprule
\textbf{Query}:\\
ROS2 --- modify Python launchfile without rebuilding package. In ROS1 we could modify XML launchfiles and \texttt{roslaunch} them directly without \texttt{catkin\_make} rebuilding, but in ROS2 it seems if I modify a Python launchfile, I need to rebuild the package; otherwise \texttt{ros2 launch <package\_name> <launchfile\_name>} runs the old launchfile. Do I need to rebuild the package after each modification, or am I missing something?\\
\midrule
\textbf{Reasoning Aspect 1 (weight = 0.30)}\\
ROS\,2 launch files are packaged and installed into the package's \texttt{share} directory during build, and \texttt{ros2 launch} executes the installed copy --- not the source.\\[4pt]
\textbf{Positive Documents}\\
\textbf{Doc 1:} (\dots abbreviation \dots) ROS\,2 tutorial ``Integrating launch files into ROS\,2 packages'' shows the standard layout (\texttt{launch/} directory at package top level) and explains that launch files are installed and consumed from the package share directory after building with \texttt{ament\_python}/\texttt{ament\_cmake}. (\dots abbreviation \dots)\\
\midrule
\textbf{Reasoning Aspect 2 (weight = 0.20)}\\
By default \texttt{colcon} performs out-of-source builds, placing built artifacts in an \texttt{install/} directory peer to \texttt{src/}, so source edits do not affect what \texttt{ros2 launch} runs.\\[4pt]
\textbf{Positive Documents}\\
\textbf{Doc 1:} (\dots abbreviation \dots) A ROS\,2 workspace contains \texttt{src/}, and \texttt{colcon} creates \texttt{build/}, \texttt{install/}, and \texttt{log/} as peers; each package is installed into its own subdirectory of \texttt{install/} (no \texttt{devel} as in catkin). (\dots abbreviation \dots)\\
\midrule
\textbf{Reasoning Aspect 3 (weight = 0.30)}\\
Building with \texttt{colcon build --symlink-install} replaces the install copies with symbolic links pointing back at the source files, so subsequent edits take effect immediately without rebuilding.\\[4pt]
\textbf{Positive Documents}\\
\textbf{Doc 1:} (\dots abbreviation \dots) \texttt{colcon} \texttt{build} verb documentation describing flags such as \texttt{--build-base}, \texttt{--install-base}, \texttt{--symlink-install}, and \texttt{--merge-install}, and how install behavior is controlled. (\dots abbreviation \dots)\\
\textbf{Doc 2:} (\dots abbreviation \dots) Community Q\&A confirms that with \texttt{--symlink-install}, source code changes take effect without recompilation (for Python; not C++). (\dots abbreviation \dots)\\
\midrule
\textbf{Reasoning Aspect 4 (weight = 0.20)}\\
With \texttt{--symlink-install} enabled, modifying a launch file (or any Python file) in \texttt{src/} becomes immediately visible to \texttt{ros2 launch} without rebuilding the package.\\[4pt]
\textbf{Positive Documents}\\
\textbf{Doc 1:} (\dots abbreviation \dots) Stack Exchange answer explaining that \texttt{--symlink-install} creates symlinks in the install directory pointing at the actual source files, so launchfile changes do not require a rebuild. (\dots abbreviation \dots)\\
\bottomrule
\end{tabularx}
\caption{\ours~Robotics Example.}
\label{tab:robotics-example}
\end{table*}

\begin{table*}[!t]
\centering
\small
\begin{tabularx}{\linewidth}{@{}X@{}}
\toprule
\textbf{Query}:
Lossy compression: \texttt{uint16}$\to$\texttt{uint8}. I need a lossy compression that maps a \texttt{uint16} to a \texttt{uint8} so that the resolution loss \emph{increases with the magnitude} of the input. My current attempt is \texttt{log2\_compress(x)} = \texttt{uint8((log2(x+1)\,/\,16)\,*\,255)}, which exploits that $\log_2(1)=0$ and $\log_2(65536)=16$. This is simple but wasteful on the \texttt{uint8} side (e.g., codewords \texttt{[1,14]}, \texttt{[16,24]}, \dots\ are never used). Can someone suggest a method similar to \texttt{log2\_compress} but that uses (more of) all 256 \texttt{uint8} levels?\\
\midrule
\textbf{Reasoning Aspect 1 (weight = 0.38)}\\
Mapping \texttt{uint16} into \texttt{uint8} with magnitude-dependent loss corresponds to \emph{nonuniform quantization} (smaller intervals where the signal is dense, larger where it is sparse), not uniform linear quantization.\\[4pt]
\textbf{Positive Documents}\\
\textbf{Doc 1:} (...abbreviation...) Textbook treatment of nonuniform quantization: by analogy with assigning shorter codewords to high-probability symbols in lossless compression, a quantizer can use smaller intervals in regions of high probability mass and larger intervals where the signal is rare, lowering average distortion. (...abbreviation...)\\
\midrule
\textbf{Reasoning Aspect 2 (weight = 0.38)}\\
$\mu$-law companding applies a logarithmic compression that concentrates quantization resolution near small magnitudes while compressing larger values --- the standard 8-bit companding scheme used in G.711 PCM telephony.\\[4pt]
\textbf{Positive Documents}\\
\textbf{Doc 1:} (...abbreviation...) The $\mu$-law algorithm (G.711) is a companding algorithm primarily used in 8-bit PCM digital telecommunications; its non-linear quantization effectively increases the dynamic range covered for a given number of bits, at a small cost in peak SNR. (...abbreviation...)\\
\midrule
\textbf{Reasoning Aspect 3 (weight = 0.25)}\\
The \emph{Lloyd--Max} algorithm designs optimal nonuniform scalar quantizers by iteratively choosing decision thresholds and reconstruction levels to minimize mean-squared quantization error for a given source PDF.\\[4pt]
\textbf{Positive Documents}\\
\textbf{Doc 1:} (...abbreviation...) Lloyd--Max scalar quantizer: for a signal $x$ with PDF $f_X(x)$ and $M$ representation levels, decision thresholds lie midway between adjacent reconstruction levels, and reconstruction levels lie at the centroid of the PDF between thresholds; the iterative procedure yields the MSE-optimal nonuniform quantizer. (...abbreviation...)\\
\bottomrule
\end{tabularx}
\caption{\ours~Stackoverflow Example.}
\label{tab:stackoverflow-example}
\end{table*}

\begin{table*}[!t]
\centering
\small
\begin{tabularx}{\linewidth}{@{}X@{}}
\toprule
\textbf{Query}:
Is it okay from a sustainability viewpoint to buy a Christmas tree? I am thinking about buying a Christmas tree but I want to know if that is or isn't good from a sustainability viewpoint.\\
\midrule

\textbf{Reasoning Aspect 1 (weight = 0.33)}\\
Life-cycle assessment (LCA) comparing natural and artificial Christmas trees generally finds that natural trees have lower annual climate and resource impacts.\\[4pt]
\textbf{Positive Documents}\\
\textbf{Doc 1:} (...abbreviation...) ISO\,14040/14044 LCA study (Montreal market) compares a plantation natural tree harvested 150\,km south of Montreal with a 6-year-reuse artificial tree manufactured in China and shipped via Vancouver; on an annual basis the artificial tree has $\sim$3$\times$ more impacts on climate change and resource depletion than the natural tree ($\sim$8.1 vs.\ $\sim$3.1 kg\,CO$_2$/yr); natural tree wins on climate change and resources but loses on ecosystem quality (mainly land occupation). (...abbreviation...)\\
\midrule

\textbf{Reasoning Aspect 2 (weight = 0.22)}\\
Artificial trees concentrate environmental impacts in manufacturing and long-distance transport, becoming climate-favorable only if reused for many years.\\[4pt]
\textbf{Positive Documents}\\
\textbf{Doc 1:} (...abbreviation...) Carbon Trust guidance: the real-vs-artificial question has no one-size-fits-all answer; ``an artificial tree, used over multiple years (7--20 years, depending on weight and materials), is better for reducing emissions than buying a new commercially grown tree every year,'' though plastic/metal components may end up as waste. (...abbreviation...)\\
\midrule

\textbf{Reasoning Aspect 3 (weight = 0.22)}\\
Sustainability of real Christmas trees depends on responsible forestry, with third-party certification systems verifying that supplying forests are managed to sustainability standards.\\[4pt]
\textbf{Positive Documents}\\
\textbf{Doc 1:} (...abbreviation...) The American Tree Farm System (ATFS), the United States' oldest family-forest certification program, applies eight sustainability standards verified by independent third-party audits; certified family forests are managed for water, wildlife, wood, and recreation. (...abbreviation...)\\
\textbf{Doc 2:} (...abbreviation...) Forest certification is defined as the certification of land-management practices to a standard of sustainability, attested by an independent third party. (...abbreviation...)\\
\midrule

\textbf{Reasoning Aspect 4 (weight = 0.22)}\\
Real Christmas trees are farm-grown over several years, can provide wildlife habitat while growing, and can be recycled into mulch or compost at end of life.\\[4pt]
\textbf{Positive Documents}\\
\textbf{Doc 1:} (...abbreviation...) Christmas tree plantations function like managed woodland meadows, providing habitat for songbirds, ground-dwelling birds, mammals, and pollinators via diverse ground covers between trees. (...abbreviation...)\\
\textbf{Doc 2:} (...abbreviation...) The U.S.\ produces Christmas trees on farms across many states (Oregon and North Carolina lead with $\sim$4.7M and $\sim$4M trees/year); typical Christmas trees are conifers with 600+ species available, grown over multiple years before harvest. (...abbreviation...)\\
\textbf{Doc 3:} (...abbreviation...) Municipal recycling programs accept natural Christmas trees for conversion into mulch and compost (decorations and flocking removed), saving landfill space; flocked trees cannot be recycled. (...abbreviation...)\\

\bottomrule
\end{tabularx}
\caption{\ours~Sustainable Living Example.}
\label{tab:sustainable-living-example}
\end{table*}

%% file: tables/appendix_metric.tex
\begin{table*}[h]
\centering
\small
\resizebox{\linewidth}{!}{
\begin{tabular}{l*{9}{C{1.27cm}}}
\toprule
\textbf{Model} &
\makecell{\textbf{\bright}\\\textbf{Overall}} &
\textbf{Biology} &
\makecell{\textbf{Earth}\\\textbf{Science}} &
\textbf{Economics} &
\textbf{Psychology} &
\textbf{Robotics} &
\makecell{\textbf{Stack}\\\textbf{Overflow}} &
\makecell{\textbf{Sustainable}\\\textbf{Living}} &
\textbf{Overall} \\
\midrule
BGE-Reasoner-8B        & 33.8   & \underline{91.0} & \textbf{91.4} & \underline{82.7} & \textbf{84.0} & \textbf{82.0} & \textbf{86.0} & \textbf{83.9} & \textbf{85.9} \\
DIVER-4B-1020          & 30.6   & \textbf{92.1} & 87.6 & \textbf{83.2} & \underline{82.2} & \underline{75.1} & \underline{83.8} & \underline{77.7} & \underline{83.1} \\
DIVER-4B               & 28.9   & 87.9 & \underline{89.6} & 74.6 & 80.9 & 69.6 & 79.3 & 74.5 & 79.5 \\
RTriever-4B (ours)     & 27.7   & 84.4 & 83.5 & 75.9 & 71.5 & 64.6 & 76.8 & 75.3 & 76.0 \\
INF-Retriever-Pro (7B) & 26.3   & 83.5 & 80.6 & 74.8 & 74.0 & 65.1 & 70.4 & 71.9 & 74.3 \\
Qwen3-8B               & 23.7   & 70.9 & 77.5 & 64.8 & 66.0 & 61.2 & 72.1 & 59.3 & 67.4 \\
\instructor-XL (1.5B)  & 18.9   & 65.9 & 74.8 & 61.3 & 66.0 & 62.1 & 65.7 & 63.8 & 65.7 \\
GTE-7B                 & 22.5   & 79.2 & 78.7 & 61.0 & 60.9 & 51.7 & 59.9 & 56.9 & 64.0 \\
OpenAI-Embed-3L        & 17.9   & 72.8 & 75.1 & 63.5 & 65.8 & 50.3 & 60.3 & 59.6 & 63.9 \\
GritLM (7B)            & 21.0   & 69.3 & 74.3 & 59.9 & 57.6 & 56.8 & 58.0 & 57.4 & 61.9 \\
ReasonIR-8B            & 24.4   & 63.6 & 69.5 & 54.9 & 54.3 & 56.4 & 68.7 & 53.4 & 60.1 \\
BM25                   & 14.5   & 58.3 & 67.6 & 57.5 & 44.6 & 53.7 & 60.6 & 59.6 & 57.4 \\
EmbeddingGemma-300M    & 18.9 & 68.6 & 76.5 & 49.5 & 51.2 & 45.9 & 53.4 & 54.7 & 57.1 \\
\bottomrule
\end{tabular}
}
\caption{A-Recall@25 across domains (\(\times 100\)). \bright Overall reports the corresponding NDCG@10 on the original \textsc{BRIGHT} dataset.}
\label{tab:model_performance_arecall}
\end{table*}

\begin{table*}[h]
\centering
\small
\resizebox{\linewidth}{!}{
\begin{tabular}{l*{9}{C{1.27cm}}}
\toprule
\textbf{Model} &
\makecell{\textbf{\bright}\\\textbf{Overall}} &
\textbf{Biology} &
\makecell{\textbf{Earth}\\\textbf{Science}} &
\textbf{Economics} &
\textbf{Psychology} &
\textbf{Robotics} &
\makecell{\textbf{Stack}\\\textbf{Overflow}} &
\makecell{\textbf{Sustainable}\\\textbf{Living}} &
\textbf{Overall} \\
\midrule
BGE-Reasoner-8B        & 33.8   & \underline{80.3} & \textbf{84.0} & \underline{71.0} & \textbf{74.4} & \textbf{71.8} & \textbf{82.1} & \textbf{68.9} & \textbf{76.1} \\
DIVER-4B-1020          & 30.6   & \textbf{81.4} & 79.0 & \textbf{71.3} & \underline{71.9} & \underline{62.1} & \underline{79.5} & \underline{62.7} & \underline{72.5} \\
DIVER-4B               & 28.9   & 77.0 & \underline{79.7} & 60.5 & 71.0 & 57.3 & 74.1 & 57.3 & 68.1 \\
RTriever-4B (ours)     & 27.7   & 69.6 & 71.1 & 60.5 & 60.7 & 53.5 & 71.5 & 57.1 & 63.4 \\
INF-Retriever-Pro (7B) & 26.3   & 70.0 & 69.2 & 59.3 & 64.9 & 53.5 & 65.5 & 53.9 & 62.3 \\
Qwen3-8B               & 23.7   & 57.2 & 64.3 & 50.0 & 56.9 & 50.4 & 66.3 & 42.5 & 55.4 \\
\instructor-XL (1.5B)  & 18.9   & 52.5 & 62.8 & 49.7 & 54.1 & 49.4 & 59.6 & 45.6 & 53.4 \\
GTE-7B                 & 22.5   & 66.1 & 68.1 & 45.3 & 49.1 & 40.5 & 53.8 & 40.4 & 51.9 \\
OpenAI-Embed-3L        & 17.9   & 56.9 & 60.1 & 48.8 & 54.7 & 38.4 & 54.9 & 42.9 & 51.0 \\
GritLM (7B)            & 21.0   & 55.2 & 62.3 & 45.2 & 48.2 & 45.6 & 52.2 & 39.9 & 49.8 \\
ReasonIR-8B            & 24.4   & 47.3 & 56.4 & 41.2 & 44.0 & 45.9 & 62.7 & 38.0 & 47.9 \\
BM25                   & 14.5   & 45.1 & 53.6 & 44.9 & 35.9 & 43.8 & 53.6 & 43.4 & 45.8 \\
EmbeddingGemma-300M    & 18.9 & 54.2 & 63.3 & 37.7 & 41.5 & 35.7 & 46.0 & 39.6 & 45.4 \\
\bottomrule
\end{tabular}
}
\caption{Recall@25 across domains (\(\times 100\)). \bright Overall reports the corresponding NDCG@10 on the original \textsc{BRIGHT} dataset.}
\label{tab:model_performance_recall}
\end{table*}

\begin{table*}[h]
\centering
\small
\resizebox{\linewidth}{!}{
\begin{tabular}{l*{9}{C{1.27cm}}}
\toprule
\textbf{Model} &
\makecell{\textbf{\bright}\\\textbf{Overall}} &
\textbf{Biology} &
\makecell{\textbf{Earth}\\\textbf{Science}} &
\textbf{Economics} &
\textbf{Psychology} &
\textbf{Robotics} &
\makecell{\textbf{Stack}\\\textbf{Overflow}} &
\makecell{\textbf{Sustainable}\\\textbf{Living}} &
\textbf{Overall} \\
\midrule
BGE-Reasoner-8B        & 33.8   & \textbf{73.3} & \textbf{75.5} & \textbf{64.1} & \textbf{64.7} & \textbf{63.0} & \textbf{68.8} & \textbf{61.7} & \textbf{67.3} \\
DIVER-4B-1020          & 30.6   & \underline{72.4} & \underline{72.9} & \underline{60.4} & 59.0 & \underline{54.0} & \underline{64.4} & \underline{56.9} & \underline{62.9} \\
DIVER-4B               & 28.9   & 67.8 & 71.5 & 51.4 & \underline{59.6} & 50.9 & 60.8 & 50.8 & 59.0 \\
RTriever-4B (ours)     & 27.7   & 61.7 & 63.4 & 51.3 & 48.6 & 46.7 & 53.4 & 50.6 & 53.7 \\
INF-Retriever-Pro (7B) & 26.3   & 60.8 & 62.8 & 48.3 & 51.3 & 44.0 & 50.3 & 47.3 & 52.1 \\
Qwen3-8B               & 23.7   & 51.2 & 57.5 & 42.8 & 45.5 & 45.6 & 52.5 & 38.2 & 47.6 \\
\instructor-XL (1.5B)  & 18.9   & 43.7 & 53.6 & 42.6 & 43.9 & 42.4 & 47.6 & 38.6 & 44.6 \\
OpenAI-Embed-3L        & 17.9   & 51.7 & 52.6 & 42.0 & 43.2 & 35.4 & 44.1 & 37.9 & 43.8 \\
GTE-7B                 & 22.5   & 58.7 & 61.4 & 35.9 & 36.2 & 39.2 & 39.7 & 34.2 & 43.6 \\
GritLM (7B)            & 21.0   & 48.7 & 56.7 & 37.5 & 36.3 & 41.3 & 38.4 & 35.8 & 42.1 \\
ReasonIR-8B            & 24.4   & 40.1 & 50.1 & 32.6 & 33.7 & 38.6 & 45.7 & 32.2 & 39.0 \\
BM25                   & 14.5   & 40.2 & 47.2 & 38.1 & 29.6 & 37.6 & 39.6 & 37.9 & 38.6 \\
EmbeddingGemma-300M    & 18.9 & 47.3 & 55.1 & 30.5 & 31.7 & 32.7 & 34.8 & 32.8 & 37.8 \\
\bottomrule
\end{tabular}
}
\caption{NDCG@25 across domains (\(\times 100\)). \bright Overall reports the corresponding NDCG@10 on the original \textsc{BRIGHT} dataset.}
\label{tab:model_performance_ndcg}
\end{table*}

%% file: tables/fix-round-qwen35.tex
\begin{table*}[!t]
\centering
\small
\resizebox{\linewidth}{!}{
\begin{tabular}{l*{9}{C{1.3cm}}}
\toprule
\multirow{2}{*}{\textbf{Model}} &
\multicolumn{3}{c}{\textbf{Round 1}} &
\multicolumn{3}{c}{\textbf{Round 2}} &
\multicolumn{3}{c}{\textbf{Round 3}} \\
\cmidrule(lr){2-4}\cmidrule(lr){5-7}\cmidrule(lr){8-10}
& \makecell{\textbf{$\bm{\alpha}$-nDCG}\\(0-100)} & \makecell{\textbf{Compl.}\\(1-5)} & \makecell{\textbf{Overall}\\(1-5)}
& \makecell{\textbf{$\bm{\alpha}$-nDCG}\\(0-100)} & \makecell{\textbf{Compl.}\\(1-5)} & \makecell{\textbf{Overall}\\(1-5)}
& \makecell{\textbf{$\bm{\alpha}$-nDCG}\\(0-100)} & \makecell{\textbf{Compl.}\\(1-5)} & \makecell{\textbf{Overall}\\(1-5)} \\
\midrule
BGE-Reasoner-8B       & \textbf{56.36} & \textbf{3.72} & \textbf{3.49} & \textbf{59.50} & \textbf{3.99} & \textbf{3.77} & \textbf{61.85} & \textbf{4.08} & \textbf{3.89} \\
GTE-7B                & 41.86 & 3.36 & 3.17 & 46.69 & 3.76 & 3.61 & \underline{49.78} & 3.89 & \underline{3.72} \\
RTriever-4B (ours)    & \underline{43.92} & \underline{3.45} & \underline{3.26} & \underline{46.94} & \underline{3.81} & \underline{3.63} & 49.65 & 3.93 & 3.71 \\
DIVER-4B              & 40.02 & 3.33 & 3.19 & 44.60 & 3.78 & 3.56 & 47.93 & \underline{3.94} & \underline{3.73} \\
Qwen3-8B              & 38.74 & 3.23 & 3.06 & 43.28 & 3.69 & 3.45 & 46.46 & 3.86 & 3.64 \\
DIVER-4B-1020         & 36.11 & 3.17 & 2.98 & 40.52 & 3.65 & 3.43 & 43.55 & 3.88 & 3.66 \\
GritLM (7B)           & 34.40 & 3.12 & 3.01 & 38.23 & 3.55 & 3.41 & 41.89 & 3.85 & 3.58 \\
BM25                  & 32.98 & 3.08 & 2.95 & 38.51 & 3.67 & 3.48 & 41.31 & 3.87 & 3.66 \\
\instructor-XL (1.5B) & 29.60 & 3.10 & 2.88 & 35.86 & 3.53 & 3.33 & 38.65 & 3.74 & 3.52 \\
ReasonIR-8B           & 30.98 & 3.12 & 2.98 & 35.57 & 3.47 & 3.29 & 37.97 & 3.65 & 3.43 \\
\bottomrule
\end{tabular}
}
\caption{Performance of retrievers under \ours fixed-round agentic evaluation setting with the Qwen3.5-122B-A10B agent (counterpart of Table~\ref{tab:roundwise}, which uses GPT-5-mini). Because each round returns top-5 passages, we report $\alpha$-nDCG at the cumulative cut-off: $k{=}5$ after round 1, $k{=}10$ after round 2, and $k{=}15$ after round 3. \textbf{Compl}: reasoning completeness of the final response; \textbf{Overall}: overall quality of the final response. Rows are ordered by round-3 $\alpha$-nDCG@15 (descending).}
\label{tab:roundwise_qwen35}
\end{table*}

%% file: appendix/case_study.tex
\begin{figure*}[t]
\centering
\begin{tcolorbox}[breakable, colback=black!3, colframe=black!70,
  title=Case Study (Early-Round Efficiency): Why is the Antarctic ice sheet only a few kilometres thick?]
\small
\setlength{\tabcolsep}{6pt}

\begin{minipage}[t]{0.56\linewidth}
\raggedright
\textbf{Aspect coverage by retrieval (3 rounds total)}
\vspace{4pt}

\begin{tabularx}{\linewidth}{@{}>{\raggedright\arraybackslash}p{0.74\linewidth} >{\raggedleft\arraybackslash}p{0.18\linewidth}@{}}
\toprule
\textbf{Aspect} (weight) & \textbf{Gold retrieved} \\
\midrule
\textbf{Outward gravitational flow} (3): snowfall accumulates in the interior but ice continually flows outward toward the coast. & 2 / 3 \\
\addlinespace[3pt]
\textbf{Self-limiting basal stress} (2): increasing thickness raises basal stress and accelerates deformation, so ice cannot grow indefinitely. & 2 / 2 \\
\addlinespace[3pt]
\textbf{Mass-balance closure} (3): interior accumulation is offset by ablation, basal melting under shelves, calving, and peripheral discharge. & 4 / 5 \\
\addlinespace[3pt]
\textbf{Slow-flow domes} (2): near central domes and divides ice velocities are very small, allowing ancient ice and subglacial features to persist. & 3 / 3 \\
\bottomrule
\end{tabularx}
\end{minipage}\hfill
\begin{minipage}[t]{0.41\linewidth}
\raggedright
\textbf{Per-round search queries issued by the agent} \\
\vspace{2pt}
\begin{itemize}[leftmargin=1em,itemsep=0pt]
    \item \textbf{R1.} ``Dome Fuji 25 mm/yr accumulation \dots oldest ice ice flow basal melting ice divides compression densification''
    \item \textbf{R2.} ``Why Antarctic ice thickness limited 3000 m equilibrium between accumulation and ice flow basal melt geothermal heat\dots''
    \item \textbf{R3.} ``Nature 1.5 million year ice core Antarctica oldest ice Vostok Dome Fuji''
\end{itemize}
\vspace{2pt}
\textbf{Final retrieval:} 11 / 13 gold; \textbf{judge:} weighted aspect coverage = 1.00, overall quality = 5.

\end{minipage}

\vspace{8pt}
\begin{tcolorbox}[colback=white, colframe=black!40, sharp corners, title={\model\ Response (excerpt)}, fonttitle=\bfseries, breakable]
\small
\textit{``Antarctic snow does not simply pile up indefinitely in one place; most of the mass that falls in the interior is continuously redistributed and removed by ice flow, basal melting/freezing and surface ablation, and the ice sheet finds a dynamic equilibrium thickness (kilometres, not hundreds of kilometres) set by those processes and by ice rheology\dots\ Ice sheets are not static piles: they flow outward from high domes toward the coast; ice at the summit is fed by snowfall but the ice mass is transported laterally by deformation and sliding so net thickness is controlled by the balance between input (snowfall) and output (flow, basal melt, calving, sublimation)\dots\ Thicker ice and increased surface slope raise driving stress, which increases deformation and flow; glaciers and ice sheets therefore thicken until flow (and other losses) balances accumulation, producing steady-state thicknesses of a few kilometres rather than hundreds of kilometres.''}
\end{tcolorbox}

\vspace{6pt}
\hrule height0.6pt
\vspace{6pt}
\noindent\textbf{What the evidence shows.}
The first two rounds retrieve ten of the eleven gold documents, covering all four reasoning aspects, and round~3 only adds a peripheral hit on slow-flow domes; the agent then terminates. Because high-quality evidence is concentrated in the first two queries, the LLM never has to speculate about why ice does not pile up indefinitely. It composes a mass-balance answer that cites the gold passages on outward flow, basal stress, ablation, and dome dynamics directly. This is the canonical pattern \model\ shows on broadly-aspected scientific queries: rapid early concentration of gold evidence, mass-balance reasoning paid out aspect-by-aspect, and a tight three-round termination.

\end{tcolorbox}
\caption{\model\ on \textit{``Why is the Antarctic ice sheet only a few kilometres thick?''} (\textsc{earth\_science} qid=48). Three search rounds suffice to cover all four reasoning aspects.}
\label{tab:case-early-round}
\end{figure*}

\begin{figure*}[t]
\centering
\begin{tcolorbox}[breakable, colback=black!3, colframe=black!70,
  title=Case Study (Evidence Deprivation): ``Can't find Apply Force Torque plugin in Gazebo Garden.'']
\small
\setlength{\tabcolsep}{6pt}

\begin{minipage}[t]{0.56\linewidth}
\raggedright
\textbf{Retrieval landscape after 13 adaptive rounds}
\vspace{4pt}

\begin{tabularx}{\linewidth}{@{}>{\raggedright\arraybackslash}p{0.66\linewidth} >{\raggedleft\arraybackslash}p{0.26\linewidth}@{}}
\toprule
\textbf{Aspect} (weight, gold doc) & \textbf{Gold retrieved} \\
\midrule
\textbf{Plugins toolbar entry point} (3, \texttt{robotics-87/extraction\_0}): GUI plugins are loaded from the toolbar Plugins button, which lists every plugin in the install. & \textbf{0 / 1} \\
\addlinespace[3pt]
\textbf{Apply Force Torque is a GUI plugin} (3, \texttt{robotics-87/extraction\_1}): the tool exists in Garden and must be selected from the plugin dropdown. & \textbf{0 / 1} \\
\bottomrule
\end{tabularx}

\vspace{6pt}
\textbf{What the agent kept retrieving instead} (24 unique non-gold docs across 13 rounds): \texttt{source\_install/tutorialstutapplyfor\_*} (Gazebo Classic tutorial), \texttt{camera\_plugin/applyjointforcesdf\_*} (the unrelated \texttt{ApplyJointForce} system), \texttt{gazebo\_detach/namespaceignition11g\_*} (the \texttt{ExternalWorldWrenchCmd} component).
\end{minipage}\hfill
\begin{minipage}[t]{0.41\linewidth}
\raggedright
\textbf{Selected search queries (paraphrased)} \\
\vspace{2pt}
\begin{itemize}[leftmargin=1em,itemsep=0pt]
    \item \textbf{R1.} ``ApplyForceTorque plugin gazebo garden''
    \item \textbf{R3.} ``Gazebo Garden ApplyForceTorque plugin"
    \item \textbf{R6.} ``Gazebo Garden plugins list documentation''
    \item \textbf{R9.} ``Ignition Gazebo apply joint force system demo''
    \item \textbf{R11.} ``Gazebo Classic vs Garden plugins compatibility''
    \item \textbf{R12.} ``ignition gazebo apply force link wrench garden''
\end{itemize}
\vspace{4pt}
\textbf{Outcome:} 0 / 2 gold retrieved; \textbf{judge:} weighted aspect coverage = 0.00, overall quality = 1.

\end{minipage}

\vspace{8pt}
\begin{tcolorbox}[colback=white, colframe=black!40, sharp corners, title={\model\ Response (excerpt), speculative}, fonttitle=\bfseries, breakable]
\small
\textit{``The Apply Force/Torque plugin referenced in the classic Gazebo tutorials is from the older (Gazebo Classic) plugin system and \emph{does not exist in the same form in Gazebo `Garden'}\dots\ In Garden the simulator uses an entity-component / system architecture instead of the old per-model plugin binary. Equivalent functionality is provided by systems/components such as an ApplyJointForce system\dots\ and by components for applying wrenches to links (ExternalWorldWrenchCmd uses an ignition msgs::Wrench) rather than the classic ApplyForceTorque plugin\dots\ \emph{So: you do not need to ``update'' Garden to get the old plugin; that plugin is simply not part of Garden.''}}
\end{tcolorbox}

\vspace{6pt}
\hrule height0.6pt
\vspace{6pt}
\noindent\textbf{What the evidence shows.} The correct answer (the plugin does ship in Garden and is loaded from the GUI's \emph{Plugins} dropdown in the toolbar) is two clicks away in the gold passages, but \model\ never surfaces them across thirteen rounds; instead it pivots through Gazebo Classic tutorials and unrelated systems (\texttt{ApplyJointForce}, \texttt{ExternalWorldWrenchCmd}). With no gold evidence to anchor it, the LLM stitches a plausible-sounding but \emph{incorrect} narrative: it concludes the plugin is deprecated and recommends programmatic SDF systems as ``equivalents.'' This is the canonical evidence-deprivation failure: the question has a simple answer, retrieval misses it, and the agent confabulates a confident substitute that contradicts the actual GUI behavior.

\end{tcolorbox}
\caption{\model\ on \textit{``Can't find Apply Force Torque plugin in Gazebo Garden''} (\textsc{robotics} qid=87). Thirteen rounds, zero gold retrieved; the model speculates that the plugin no longer exists.}
\label{tab:case-evidence-deprivation}
\end{figure*}

\begin{figure*}[t]
\centering
\begin{tcolorbox}[breakable, colback=black!3, colframe=black!70,
  title=Case Study (Repetition Bias): Has interfamilial hybridization ever succeeded?]
\small
\setlength{\tabcolsep}{6pt}

\begin{minipage}[t]{0.56\linewidth}
\raggedright
\textbf{Documents repeatedly resurfaced across 12 rounds} \\
\vspace{4pt}
\begin{tabularx}{\linewidth}{@{}>{\raggedright\arraybackslash}p{0.58\linewidth} >{\raggedright\arraybackslash}p{0.18\linewidth} >{\raggedleft\arraybackslash}p{0.10\linewidth}@{}}
\toprule
\textbf{Document ID} & \textbf{Type} & \textbf{Times} \\
\midrule
\texttt{Primate\_3\_14} (closest non-primate relatives) & non-gold / off-topic & 7 \\
\texttt{Human\_81\_1} (human monogamy) & non-gold / off-topic & 5 \\
\texttt{Human\_14\_483} (human monogamy) & non-gold / off-topic & 3 \\
\texttt{Human\_14\_127} (human monogamy) & non-gold / off-topic & 3 \\
\texttt{biology-64d3516332ed} & non-gold / off-topic & 3 \\
\texttt{Parthenogenesis\_4\_5} & non-gold / tangential & 3 \\
\texttt{Parthenogenesis\_4\_2} & non-gold / tangential & 3 \\
\addlinespace[2pt]
\textbf{Subtotal (non-gold repeats)} & & \textbf{27} \\
\midrule
\texttt{biology-86/extraction\_4} & gold (1 of 7) & 4 \\
\texttt{Reproductive\_isolation\_2} & gold (1 of 7) & 3 \\
\addlinespace[2pt]
\textbf{Subtotal (gold repeats)} & & \textbf{7} \\
\bottomrule
\end{tabularx}

\vspace{4pt}
\textbf{Aggregate:} 60 retrieval slots $\rightarrow$ 28 unique documents $\rightarrow$ 32 cross-round duplicates ($\approx$53\% of the slot budget); \textbf{gold:} 3 / 7 retrieved (extraction\_4, extraction\_5, Reproductive\_isolation\_2); \textbf{judge:} weighted aspect coverage = 0.59, overall quality = 3.
\end{minipage}\hfill
\begin{minipage}[t]{0.41\linewidth}
\raggedright
\textbf{What the agent never reaches} \\
\vspace{2pt}
\begin{itemize}[leftmargin=1em,itemsep=0pt]
    \item \texttt{Reproductive\_isolation\_0/\_1} (pre-zygotic barriers; Aspect~1, weight~3)
    \item \texttt{biology-86/extraction\_0} (post-zygotic isolation; Aspect~2, weight~3)
    \item \texttt{biology-86/extraction\_3} (interfamilial fish hybrids; Aspect~3, weight~3)
\end{itemize}
\vspace{4pt}
The four \emph{Ursus} hybrid documents and the daylily--Lycoris embryo-rescue passage are reached, but only after rounds~3 and~6; the rest of the run circles back to the same primate / human-monogamy / parthenogenesis cluster.

\end{minipage}

\vspace{8pt}
\begin{tcolorbox}[colback=white, colframe=black!40, sharp corners, title={\model\ Response (excerpt), partial coverage}, fonttitle=\bfseries, breakable]
\small
\textit{``In animals: basically no credible examples \dots\ of true extra-familial sexual hybrids that produced viable, reproducing offspring; what is documented are interspecific and intergeneric hybrids within the same family or closely related genera\dots\ In plants: extra-familial hybrids are rare but have been produced under artificial (laboratory/breeding) conditions using assisted methods such as embryo rescue; at least one documented case produced a plantlet identified as an interfamily hybrid\dots\ A reported study crossed daylily (\emph{Hemerocallis}) with \emph{Lycoris} and, after many aborted ovaries and using assisted techniques, recovered one embryo that developed into a plantlet identified as an interfamily hybrid (\emph{L.~aurea} $\times$ \emph{H.~`My Complementary'})\dots\ The literature emphasizes that hybrid formation yielding fertile, stable offspring is uncommon in animals because post-zygotic barriers (chromosomal mismatches, gene incompatibilities) prevent viable, fertile hybrids as genetic distance increases.''}
\end{tcolorbox}

\vspace{6pt}
\hrule height0.6pt
\vspace{6pt}
\noindent\textbf{What the evidence shows.} The first round latches onto \texttt{Primate\_3\_14} and the parthenogenesis cluster; over the next eleven rounds the agent rephrases its query seven different ways (``intergeneric hybridization,'' ``hybrids between different families,'' ``somatic hybridization protoplast fusion,'' ``gelada hamadryas baboon hybrid,'' \dots) but the retriever resurfaces the same anchors plus an unrelated human-monogamy cluster, producing 32 cross-round duplicate retrievals on a 60-slot budget. The agent does eventually find three of the seven gold documents, but the two highest-weighted aspects (\emph{pre-zygotic barriers} and \emph{post-zygotic isolation}, both weight~3) are never reached. The final response correctly describes the daylily--Lycoris example but treats reproductive-isolation mechanisms only at a generic level, yielding the partial answer that the aspect-aware judge scores at $\text{wac}=0.59$. The failure mode is search-dynamics: feedback from the early rounds narrows rather than expands the candidate set, even when the agent is actively trying to broaden the query.

\end{tcolorbox}
\caption{\model\ on \textit{``Is sexual reproduction outside the same biological family possible?''} (\textsc{biology} qid=86). Repeated off-target retrievals dominate twelve adaptive rounds.}
\label{tab:case-repetition}
\end{figure*}

\begin{figure*}[t]
\centering
\begin{tcolorbox}[breakable, colback=black!3, colframe=black!70,
  title={Case Study (Aspect Tunnel Vision): GHCN climate data; what is EMXT 317, and how do I get the bulk data?}]
\small
\setlength{\tabcolsep}{6pt}

\begin{minipage}[t]{0.56\linewidth}
\raggedright
\textbf{Per-aspect retrieval coverage (7 adaptive rounds)} \\
\vspace{4pt}
\begin{tabularx}{\linewidth}{@{}>{\raggedright\arraybackslash}p{0.86\linewidth} >{\raggedleft\arraybackslash}p{0.08\linewidth}@{}}
\toprule
\textbf{Aspect} (weight, gold doc) & \textbf{Retr.}\\
\midrule
\textbf{a1.} EMXT denotes extreme maximum temperature, derived from daily TMAX (\texttt{ncdc-meta}). & \cmark \\
\addlinespace[2pt]
\textbf{a2.} Climate-division datasets provide bulk monthly temperature/precipitation by division/county, often in CSV (\texttt{earth-sci-98/ext\_0,\_1}). & \xmark \\
\addlinespace[2pt]
\textbf{a3.} The NCEI Access Data Service API enables HTTP queries returning CSV (\texttt{earth-sci-98/ext\_2}). & \xmark \\
\bottomrule
\end{tabularx}

\vspace{4pt}
The user explicitly asked two things: (i) what does the value 317 mean, and (ii) is there a better way to get U.S.~monthly climate data from 1992--2012. The agent saturates on (i) and never queries any concept relevant to (ii).
\end{minipage}\hfill
\begin{minipage}[t]{0.41\linewidth}
\raggedright
\textbf{All seven search queries (paraphrased)} \\
\vspace{2pt}
\begin{itemize}[leftmargin=1em,itemsep=1pt]
    \item \textbf{R1.} GHCN monthly EMXT meaning, code, units, ``317''
    \item \textbf{R2.} GHCN Monthly CSV units tenths degrees EMXT
    \item \textbf{R3.} GHCN-Daily TMAX units tenths format
    \item \textbf{R4.} GHCN-Monthly EMXT 317 interpretation, scale
    \item \textbf{R5.} GHCN-Daily ``tenths of degrees,'' TMAX
    \item \textbf{R6.} EMXT CSV sample, Petersburg ND
    \item \textbf{R7.} ghcnd file format TMAX tenths
\end{itemize}
\vspace{4pt}
\textbf{Judge:} aspect a1 score = 1.00; a2 = a3 = 0; weighted aspect coverage = 0.50, overall quality = 3.
\end{minipage}

\vspace{8pt}
\begin{tcolorbox}[colback=white, colframe=black!40, sharp corners, title={\model\ Response (excerpt), detailed but lop-sided}, fonttitle=\bfseries, breakable]
\small
\textit{``EMXT is the `extreme maximum temperature' for that station-month, \ie\ the single highest daily maximum temperature within the month\dots\ The CSV/text export is the route to get machine-readable, multi-station data\dots\ How you can get the data you need (state or county, 1992--2012) in CSV form: \emph{(1) Use the GHCN Monthly Summaries Custom ASCII Spreadsheet (CSV) export}\dots\ \emph{(2) The Monthly Summaries are derived from the GHCN-Daily data; if you need different aggregation rules you can download GHCN-Daily and compute monthly/areal aggregates yourself}\dots''}
\end{tcolorbox}

\vspace{6pt}
\hrule height0.6pt
\vspace{6pt}
\noindent\textbf{What the evidence shows.} Every one of the seven queries elaborates the same sub-question (``what does EMXT mean / what units''). The user's other half (which dataset to use for bulk 1992--2012 county-level data) is never queried, so the climate-division dataset and the NCEI Access Data Service API are entirely absent from the retrieval set. The final answer is technically correct on the EMXT semantics and even produces a partial workflow recommendation, but it improvises that workflow on top of the GHCN-Monthly Summaries product (the only one the agent has retrieved evidence for) rather than directing the user to the dataset family designed for their use case. Aspect-aware evaluation surfaces the failure: $a_2$ and $a_3$ receive zero coverage, and weighted aspect coverage drops to $0.50$ despite a fluent, well-cited response. The pattern of fluent answer combined with unbalanced retrieval is distinct from repetition bias because the search queries themselves \emph{are} novel each round; the lock-in is semantic, not document-level.

\end{tcolorbox}
\caption{\model\ on a two-part GHCN climate-data question (\textsc{earth\_science} qid=98). All seven search rounds drill into one aspect; the user's data-source question is never queried.}
\label{tab:case-aspect-tunnel}
\end{figure*}

\begin{figure*}[t]
\centering
\begin{tcolorbox}[breakable, colback=black!3, colframe=black!70,
  title=Case Study (Hypothesis Hopping): Is there a name for ``perceived repeated interruption''?]
\small
\setlength{\tabcolsep}{6pt}

\begin{minipage}[t]{0.56\linewidth}
\raggedright
\textbf{Round-by-round behaviour} \\
\vspace{4pt}
\begin{tabularx}{\linewidth}{@{}>{\raggedright\arraybackslash}p{0.16\linewidth} >{\raggedright\arraybackslash}p{0.50\linewidth} >{\raggedright\arraybackslash}p{0.30\linewidth}@{}}
\toprule
\textbf{Round} & \textbf{Search hypothesis} & \textbf{New gold (and aspect)} \\
\midrule
R1 & ``perceived repeated interruption / single stall'' & \texttt{ext\_1} \;(a1 frequency illusion) \\
R2 & ``Baader--Meinhof phenomenon, confirmation bias'' & \texttt{ext\_4} (a4), \texttt{ext\_6} (a5) \\
R3 & spotlight effect, heightened self-awareness & none (non-gold) \\
R4 & illusion of control, ``being followed'' & none (non-gold) \\
R5 & Jungian synchronicity, meaningful coincidences & none (re-hits \texttt{ext\_6}) \\
R6 & clustering illusion, patterns in random events & \texttt{ext\_5} \;(a4) \\
\bottomrule
\end{tabularx}

\vspace{4pt}
After round~2 the agent already has a directly-named answer (the \emph{frequency illusion / Baader--Meinhof phenomenon}), plus supporting evidence on availability and patternicity. Rounds~3--6 enumerate alternative concept labels rather than terminating.
\end{minipage}\hfill
\begin{minipage}[t]{0.41\linewidth}
\raggedright
\textbf{Outcome statistics} \\
\vspace{2pt}
\begin{itemize}[leftmargin=1em,itemsep=0pt]
    \item Rounds: 6 (vs.\ task median 5)
    \item Gold retrieved: 4 / 8 (\texttt{ext\_1, 4, 5, 6})
    \item First gold round: 1; last \emph{new} gold round: 6
    \item Aspect scores: $a_1\!=\!a_2\!=\!a_3\!=\!a_4\!=\!1$, $a_5\!=\!0.5$
    \item Weighted aspect coverage = 0.95, overall quality = 4
\end{itemize}
\vspace{4pt}
The retrieved set ends up dominated by 5 \texttt{spotlight\_effect} passages and 5 \texttt{illusion\_of\_control} passages, neither tied to a gold-supported aspect.

\end{minipage}

\vspace{8pt}
\begin{tcolorbox}[colback=white, colframe=black!40, sharp corners, title={\model\ Response (excerpt), correct but bloated}, fonttitle=\bfseries, breakable]
\small
\textit{``There isn't a single agreed-upon name for the exact feeling you describe, but a number of well-documented cognitive biases and concepts cover it; most directly the `frequency illusion' (also called the Baader--Meinhof phenomenon) and the general tendency to perceive patterns in randomness often labeled apophenia or patternicity\dots\ Other closely related bias names\dots\ illusory correlation and the clustering illusion\dots\ Two additional, related concepts people often invoke: the spotlight effect (feeling that others are paying more attention to you than they actually are) can increase self-consciousness about interruptions but is conceptually different\dots\ Jungian `synchronicity' is a non-empirical label\dots\ The illusion of control can also play a role\dots''}
\end{tcolorbox}

\vspace{6pt}
\hrule height0.6pt
\vspace{6pt}
\noindent\textbf{What the evidence shows.} Unlike the early-round case where convergence is rewarded, here \emph{the agent already has the answer} after round~2 but cannot stop. Each subsequent round tests a fresh hypothesis (\textit{spotlight effect, illusion of control, synchronicity, clustering illusion}) before terminating. The final answer remains correct (judge oq~$=$~4, weighted aspect coverage~$=$~0.95), but the retrieval set is diluted with ten passages on the spotlight effect and illusion of control that play no causal role in the answer, and the response itself reads as a survey rather than a direct identification. This pattern is the converse of evidence deprivation: retrieval succeeds early, but lacking a confidence signal, the agent keeps spending rounds. From a deployment standpoint the cost is real, since each extra round adds an LLM call and a retrieval pass, yet it is invisible to final-round-only metrics.

\end{tcolorbox}
\caption{\model\ on \textit{``Is there a name for `perceived repeated interruption'?''} (\textsc{psychology} qid=98). Gold answer is retrieved by round~2; rounds~3--6 enumerate alternative concept names.}
\label{tab:case-hypothesis-hopping}
\end{figure*}

%% file: custom.bib
@inproceedings{
su2025bright,
title={{BRIGHT}: A Realistic and Challenging Benchmark for Reasoning-Intensive Retrieval},
author={Hongjin SU and Howard Yen and Mengzhou Xia and Weijia Shi and Niklas Muennighoff and Han-yu Wang and Liu Haisu and Quan Shi and Zachary S Siegel and Michael Tang and Ruoxi Sun and Jinsung Yoon and Sercan O Arik and Danqi Chen and Tao Yu},
booktitle={The Thirteenth International Conference on Learning Representations},
year={2025},
url={https://openreview.net/forum?id=ykuc5q381b}
}

@article{chen2025browsecomp,
  title={Browsecomp-plus: A more fair and transparent evaluation benchmark of deep-research agent},
  author={Chen, Zijian and Ma, Xueguang and Zhuang, Shengyao and Nie, Ping and Zou, Kai and Liu, Andrew and Green, Joshua and Patel, Kshama and Meng, Ruoxi and Su, Mingyi and others},
  journal={arXiv preprint arXiv:2508.06600},
  year={2025}
}

@misc{litsearch,
      title={LitSearch: A Retrieval Benchmark for Scientific Literature Search}, 
      author={Anirudh Ajith and Mengzhou Xia and Alexis Chevalier and Tanya Goyal and Danqi Chen and Tianyu Gao},
      year={2024},
      eprint={2407.18940},
      archivePrefix={arXiv},
      primaryClass={cs.IR},
      url={https://arxiv.org/abs/2407.18940}, 
}

@misc{bright,
  title={BRIGHT: A Realistic and Challenging Benchmark for Reasoning-Intensive Retrieval},
  author={Su, Hongjin and Yen, Howard and Xia, Mengzhou and Shi, Weijia and Muennighoff, Niklas and Wang, Han-yu and Liu, Haisu and Shi, Quan and Siegel, Zachary S and Tang, Michael and Sun, Ruoxi and Yoon, Jinsung and Arik, Sercan O and Chen, Danqi and Yu, Tao},
  url={https://arxiv.org/abs/2407.12883},
  year={2024},
}

@article{rar-b,
  title={RAR-b: Reasoning as Retrieval Benchmark},
  author={Xiao, Chenghao and Hudson, G Thomas and Moubayed, Noura Al},
  journal={arXiv preprint arXiv:2404.06347},
  year={2024}
}

@inproceedings{beir,
    author = {Kamalloo, Ehsan and Thakur, Nandan and Lassance, Carlos and Ma, Xueguang and Yang, Jheng-Hong and Lin, Jimmy},
    title = {Resources for Brewing BEIR: Reproducible Reference Models and Statistical Analyses},
    year = {2024},
    url = {https://doi.org/10.1145/3626772.3657862},
    booktitle = {Proceedings of the 47th International ACM SIGIR Conference on Research and Development in Information Retrieval},
    pages = {1431–1440},
}

@misc{scholarqa,
      title={Ai2 Scholar QA: Organized Literature Synthesis with Attribution}, 
      author={Amanpreet Singh and Joseph Chee Chang and Chloe Anastasiades and Dany Haddad and Aakanksha Naik and Amber Tanaka and Angele Zamarron and Cecile Nguyen and Jena D. Hwang and Jason Dunkleberger and Matt Latzke and Smita Rao and Jaron Lochner and Rob Evans and Rodney Kinney and Daniel S. Weld and Doug Downey and Sergey Feldman},
      year={2025},
      eprint={2504.10861},
      archivePrefix={arXiv},
      primaryClass={cs.CL},
      url={https://arxiv.org/abs/2504.10861}, 
}

@inproceedings{
shao2025reasonir,
title={Reason{IR}: Training Retrievers for Reasoning Tasks},
author={Rulin Shao and Rui Qiao and Varsha Kishore and Niklas Muennighoff and Xi Victoria Lin and Daniela Rus and Bryan Kian Hsiang Low and Sewon Min and Wen-tau Yih and Pang Wei Koh and Luke Zettlemoyer},
booktitle={Second Conference on Language Modeling},
year={2025},
url={https://openreview.net/forum?id=kkBCNLMbGj}
}

@misc{weller2025rank1testtimecomputereranking,
      title={Rank1: Test-Time Compute for Reranking in Information Retrieval}, 
      author={Orion Weller and Kathryn Ricci and Eugene Yang and Andrew Yates and Dawn Lawrie and Benjamin Van Durme},
      year={2025},
      eprint={2502.18418},
      archivePrefix={arXiv},
      primaryClass={cs.IR},
      url={https://arxiv.org/abs/2502.18418}, 
}

@misc{long2025diver,
      title={DIVER: A Multi-Stage Approach for Reasoning-intensive Information Retrieval}, 
      author={Meixiu Long and Duolin Sun and Dan Yang and Junjie Wang and Yue Shen and Jian Wang and Peng Wei and Jinjie Gu and Jiahai Wang},
      year={2025},
      eprint={2508.07995},
      archivePrefix={arXiv},
      primaryClass={cs.IR},
      url={https://arxiv.org/abs/2508.07995}, 
}

@misc{das2025rader,
      title={RaDeR: Reasoning-aware Dense Retrieval Models}, 
      author={Debrup Das and Sam O' Nuallain and Razieh Rahimi},
      year={2025},
      eprint={2505.18405},
      archivePrefix={arXiv},
      primaryClass={cs.CL},
      url={https://arxiv.org/abs/2505.18405}, 
}

@misc{bge_reasoner,
  title        = {{BGE-Reasoner}: Reasoning-aware Retrieval Framework},
  author       = {USTC and BAAI and FlagOpen Team},
  year         = {2025},
  howpublished = {\url{https://github.com/FlagOpen/FlagEmbedding}},
  note         = {Accessed: 2025-10-04}
}

@article{zhang2025diffusion,
  title={Diffusion vs. autoregressive language models: A text embedding perspective},
  author={Zhang, Siyue and Zhao, Yilun and Geng, Liyuan and Cohan, Arman and Luu, Anh Tuan and Zhao, Chen},
  journal={arXiv preprint arXiv:2505.15045},
  year={2025}
}

@article{promptriever,
  title={Promptriever: Instruction-Trained Retrievers Can Be Prompted Like Language Models},
  author={Weller, Orion and Van Durme, Benjamin and Lawrie, Dawn and Paranjape, Ashwin and Zhang, Yuhao and Hessel, Jack},
  journal={arXiv preprint arXiv:2409.11136},
  year={2024}
}

@article{nv-embed,
  title={NV-Embed: Improved Techniques for Training LLMs as Generalist Embedding Models},
  author={Lee, Chankyu and Roy, Rajarshi and Xu, Mengyao and Raiman, Jonathan and Shoeybi, Mohammad and Catanzaro, Bryan and Ping, Wei},
  journal={arXiv preprint arXiv:2405.17428},
  year={2024}
}

@inproceedings{instructor,
  title={One Embedder, Any Task: Instruction-Finetuned Text Embeddings},
  author={Su, Hongjin and Shi, Weijia and Kasai, Jungo and Wang, Yizhong and Hu, Yushi and Ostendorf, Mari and Yih, Wen-tau and Smith, Noah A and Zettlemoyer, Luke and Yu, Tao},
  booktitle={Findings of the Association for Computational Linguistics: ACL 2023},
  pages={1102--1121},
  year={2023}
}

@article{gte,
  title={Towards general text embeddings with multi-stage contrastive learning},
  author={Li, Zehan and Zhang, Xin and Zhang, Yanzhao and Long, Dingkun and Xie, Pengjun and Zhang, Meishan},
  journal={arXiv preprint arXiv:2308.03281},
  year={2023}
}

@article{qwen3embed,
  title={Qwen3 Embedding: Advancing Text Embedding and Reranking Through Foundation Models},
  author={Zhang, Yanzhao and Li, Mingxin and Long, Dingkun and Zhang, Xin and Lin, Huan and Yang, Baosong and Xie, Pengjun and Yang, An and Liu, Dayiheng and Lin, Junyang and others},
  journal={arXiv preprint arXiv:2506.05176},
  year={2025}
}

@inproceedings{andcg,
author = {Clarke, Charles L.A. and Kolla, Maheedhar and Cormack, Gordon V. and Vechtomova, Olga and Ashkan, Azin and B\"{u}ttcher, Stefan and MacKinnon, Ian},
title = {Novelty and diversity in information retrieval evaluation},
year = {2008},
isbn = {9781605581644},
publisher = {Association for Computing Machinery},
address = {New York, NY, USA},
url = {https://doi.org/10.1145/1390334.1390446},
doi = {10.1145/1390334.1390446},
booktitle = {Proceedings of the 31st Annual International ACM SIGIR Conference on Research and Development in Information Retrieval},
pages = {659–666},
numpages = {8},
location = {Singapore, Singapore},
series = {SIGIR '08}
}

@article{robertson2009probabilistic,
  title={The probabilistic relevance framework: BM25 and beyond},
  author={Robertson, Stephen and Zaragoza, Hugo and others},
  journal={Foundations and Trends{\textregistered} in Information Retrieval},
  volume={3},
  number={4},
  pages={333--389},
  year={2009},
  publisher={Now Publishers, Inc.}
}

@article{su2022one, title = {One embedder, any task: Instruction-finetuned text embeddings}, author = {Su, Hongjin and Shi, Weijia and Kasai, Jungo and Wang, Yizhong and Hu, Yushi and Ostendorf, Mari and Yih, Wen-tau and Smith, Noah A and Zettlemoyer, Luke and Yu, Tao}, journal = {arXiv preprint arXiv:2212.09741}, year = {2022} }

@article{muennighoff2024generative, title = {Generative representational instruction tuning}, author = {Muennighoff, Niklas and Su, Hongjin and Wang, Liang and Yang, Nan and Wei, Furu and Yu, Tao and Singh, Amanpreet and Kiela, Douwe}, journal = {arXiv preprint arXiv:2402.09906}, year = {2024} }

@article{li2023towards,
  title={Towards general text embeddings with multi-stage contrastive learning},
  author={Li, Zehan and Zhang, Xin and Zhang, Yanzhao and Long, Dingkun and Xie, Pengjun and Zhang, Meishan},
  journal={arXiv preprint arXiv:2308.03281},
  year={2023}
}

@misc{swesmith,
    title={SWE-smith: Scaling Data for Software Engineering Agents},
    author={John Yang and Kilian Lieret and Carlos E. Jimenez and Alexander Wettig and Kabir Khandpur and Yanzhe Zhang and Binyuan Hui and Ofir Press and Ludwig Schmidt and Diyi Yang},
    year={2025},
    eprint={2504.21798},
    archivePrefix={arXiv},
    primaryClass={cs.SE},
    url={https://arxiv.org/abs/2504.21798},
}

@article{webwalker,
  title={Webwalker: Benchmarking llms in web traversal},
  author={Wu, Jialong and Yin, Wenbiao and Jiang, Yong and Wang, Zhenglin and Xi, Zekun and Fang, Runnan and Zhang, Linhai and He, Yulan and Zhou, Deyu and Xie, Pengjun and others},
  journal={arXiv preprint arXiv:2501.07572},
  year={2025}
}

@article{deepresearchbench,
  title={DeepResearch Bench: A Comprehensive Benchmark for Deep Research Agents},
  author={Du, Mingxuan and Xu, Benfeng and Zhu, Chiwei and Wang, Xiaorui and Mao, Zhendong},
  journal={arXiv preprint arXiv:2506.11763},
  year={2025}
}

@misc{yang2025qwen3technicalreport,
      title={Qwen3 Technical Report}, 
      author={An Yang and Anfeng Li and Baosong Yang and Beichen Zhang and Binyuan Hui and Bo Zheng and Bowen Yu and Chang Gao and Chengen Huang and Chenxu Lv and Chujie Zheng and Dayiheng Liu and Fan Zhou and Fei Huang and Feng Hu and Hao Ge and Haoran Wei and Huan Lin and Jialong Tang and Jian Yang and Jianhong Tu and Jianwei Zhang and Jianxin Yang and Jiaxi Yang and Jing Zhou and Jingren Zhou and Junyang Lin and Kai Dang and Keqin Bao and Kexin Yang and Le Yu and Lianghao Deng and Mei Li and Mingfeng Xue and Mingze Li and Pei Zhang and Peng Wang and Qin Zhu and Rui Men and Ruize Gao and Shixuan Liu and Shuang Luo and Tianhao Li and Tianyi Tang and Wenbiao Yin and Xingzhang Ren and Xinyu Wang and Xinyu Zhang and Xuancheng Ren and Yang Fan and Yang Su and Yichang Zhang and Yinger Zhang and Yu Wan and Yuqiong Liu and Zekun Wang and Zeyu Cui and Zhenru Zhang and Zhipeng Zhou and Zihan Qiu},
      year={2025},
      eprint={2505.09388},
      archivePrefix={arXiv},
      primaryClass={cs.CL},
      url={https://arxiv.org/abs/2505.09388}, 
}

@article{qwen3embedding,
  title={Qwen3 Embedding: Advancing Text Embedding and Reranking Through Foundation Models},
  author={Zhang, Yanzhao and Li, Mingxin and Long, Dingkun and Zhang, Xin and Lin, Huan and Yang, Baosong and Xie, Pengjun and Yang, An and Liu, Dayiheng and Lin, Junyang and Huang, Fei and Zhou, Jingren},
  journal={arXiv preprint arXiv:2506.05176},
  year={2025}
}

@misc{li2024llmsasjudges,
      title={LLMs-as-Judges: A Comprehensive Survey on LLM-based Evaluation Methods}, 
      author={Haitao Li and Qian Dong and Junjie Chen and Huixue Su and Yujia Zhou and Qingyao Ai and Ziyi Ye and Yiqun Liu},
      year={2024},
      eprint={2412.05579},
      archivePrefix={arXiv},
      primaryClass={cs.CL},
      url={https://arxiv.org/abs/2412.05579}, 
}

@misc{neelakantan2022text,
      title={Text and Code Embeddings by Contrastive Pre-Training},
      author={Arvind Neelakantan and Tao Xu and Raul Puri and Alec Radford and Jesse Michael Han and Jerry Tworek and Qiming Yuan and Nikolas Tezak and Jong Wook Kim and Chris Hallacy and Johannes Heidecke and Pranav Shyam and Boris Power and Tyna Eloundou Nekoul and Girish Sastry and Gretchen Krueger and David Schnurr and Felipe Petroski Such and Kenny Hsu and Madeleine Thompson and Tabarak Khan and Toki Sherbakov and Joanne Jang and Peter Welinder and Lilian Weng},
      year={2022},
      eprint={2201.10005},
      archivePrefix={arXiv},
      primaryClass={cs.CL},
      url={https://arxiv.org/abs/2201.10005},
}

@Article{Vera2025EmbeddingGemmaPA,
 author = {Henrique Schechter Vera and Sahil Dua and Biao Zhang and Daniel M. Salz and Ryan Mullins and Sindhu Raghuram Panyam and Sara Smoot and Iftekhar Naim and Joe Zou and Feiyang Chen and Daniel Cer and Alice Lisak and Min Choi and Lucas Gonzalez and Omar Sanseviero and Glenn Cameron and Ian Ballantyne and Kat Black and Kaifeng Chen and Weiyi Wang and Zhe Li and Gus Martins and Jinhyuk Lee and Mark Sherwood and Juyeong Ji and Renjie Wu and Jing Zheng and Jyotinder Singh and Abheesht Sharma and Divya Sreepat and Aashi Jain and Adham Elarabawy and A. Co. and Andreas Doumanoglou and Babak Samari and B. Hora and B. Potetz and Dahun Kim and Enrique Alfonseca and Fedor Moiseev and Feng Han and Frank Palma Gomez and Gustavo Hernández Abrego and He Zhang and Hui Hui and Jay Han and Karan Gill and Ke Chen and Koert Chen and Madhuri Shanbhogue and Michael Boratko and P. Suganthan and Sai Meher Karthik Duddu and Sandeep Mariserla and Setareh Ariafar and Shanfeng Zhang and Shijie Zhang and Simon Baumgartner and Sonam Goenka and S. Qiu and T. Dabral and Trevor Walker and Vikram Rao and Waleed Khawaja and Wenlei Zhou and Xiaoqi Ren and Ye Xia and Yichang Chen and Yi-ting Chen and Zhe Dong and Zhongli Ding and Francesco Visin and Gael Liu and Jiageng Zhang and Kathleen Kenealy and M. Casbon and Ravin Kumar and Thomas Mesnard and Zach Gleicher and C. Brick and Olivier Lacombe and A. Roberts and Yunhsuan Sung and Raphael Hoffmann and Tris Warkentin and Armand Joulin and Tom Duerig and Mojtaba Seyedhosseini},
 booktitle = {arXiv.org},
 journal = {ArXiv},
 title = {EmbeddingGemma: Powerful and Lightweight Text Representations},
 volume = {abs/2509.20354},
 year = {2025}
}

@Article{Zhao2024SWIFTASL,
 author = {Yuze Zhao and Jintao Huang and Jinghan Hu and Xingjun Wang and Yunlin Mao and Daoze Zhang and Zeyinzi Jiang and Zhikai Wu and Baole Ai and Ang Wang and Wenmeng Zhou and Yingda Chen},
 booktitle = {AAAI Conference on Artificial Intelligence},
 journal = {ArXiv},
 title = {SWIFT:A Scalable lightWeight Infrastructure for Fine-Tuning},
 volume = {abs/2408.05517},
 year = {2024}
}

@misc {infly-ai_2025,
    author       = { Junhan Yang and Jiahe Wan and Yichen Yao and Wei Chu and Yinghui Xu and Emma Wang and Yuan Qi },
    title        = { inf-retriever-v1 (Revision 5f469d7) },
    year         = 2025,
    url          = { https://huggingface.co/infly/inf-retriever-v1 },
    doi          = { 10.57967/hf/4262 },
    publisher    = { Hugging Face }
}

@Article{Campos2016MSMA,
 author = {Daniel Fernando Campos and Tri Nguyen and Mir Rosenberg and Xia Song and Jianfeng Gao and Saurabh Tiwary and Rangan Majumder and L. Deng and Bhaskar Mitra},
 booktitle = {CoCo@NIPS},
 journal = {ArXiv},
 title = {MS MARCO: A Human Generated MAchine Reading COmprehension Dataset},
 volume = {abs/1611.09268},
 year = {2016}
}

@Article{Chan2024ScalingSD,
 author = {Xin Chan and Xiaoyang Wang and Dian Yu and Haitao Mi and Dong Yu},
 booktitle = {arXiv.org},
 journal = {ArXiv},
 title = {Scaling Synthetic Data Creation with 1,000,000,000 Personas},
 volume = {abs/2406.20094},
 year = {2024}
}

@misc{shao2025drtulu,
      title={DR Tulu: Reinforcement Learning with Evolving Rubrics for Deep Research}, 
      author={Rulin Shao and Akari Asai and Shannon Zejiang Shen and Hamish Ivison and Varsha Kishore and Jingming Zhuo and Xinran Zhao and Molly Park and Samuel G. Finlayson and David Sontag and Tyler Murray and Sewon Min and Pradeep Dasigi and Luca Soldaini and Faeze Brahman and Wen-tau Yih and Tongshuang Wu and Luke Zettlemoyer and Yoon Kim and Hannaneh Hajishirzi and Pang Wei Koh},
      year={2025},
      eprint={2511.19399},
      archivePrefix={arXiv},
      primaryClass={cs.CL},
      url={https://arxiv.org/abs/2511.19399}, 
}

@misc{yifei2025researchqa,
      title={ResearchQA: Evaluating Scholarly Question Answering at Scale Across 75 Fields with Survey-Mined Questions and Rubrics}, 
      author={Li S. Yifei and Allen Chang and Chaitanya Malaviya and Mark Yatskar},
      year={2025},
      eprint={2509.00496},
      archivePrefix={arXiv},
      primaryClass={cs.CL},
      url={https://arxiv.org/abs/2509.00496}, 
}

@misc{wang2025liveresearchbench,
      title={LiveResearchBench: A Live Benchmark for User-Centric Deep Research in the Wild},
      author={Jiayu Wang and Yifei Ming and Riya Dulepet and Qinglin Chen and Austin Xu and Zixuan Ke and Frederic Sala and Aws Albarghouthi and Caiming Xiong and Shafiq Joty},
      year={2025},
      eprint={2510.14240},
      archivePrefix={arXiv},
      primaryClass={cs.AI},
      url={https://arxiv.org/abs/2510.14240},
}

@Article{Abdallah2026MMBRIGHTAM,
 author = {Abdelrahman Abdallah and Mohamed Darwish Mounis and Mahmoud Abdalla and M. Kasem and Mostafa Farouk Senussi and Mohamed Mahmoud and Mohammed Ali and Adam Jatowt and H. Kang},
 booktitle = {arXiv.org},
 journal = {ArXiv},
 title = {MM-BRIGHT: A Multi-Task Multimodal Benchmark for Reasoning-Intensive Retrieval},
 volume = {abs/2601.09562},
 year = {2026}
}

@Inproceedings{Chen2026AgentIRRR,
 author = {Zijian Chen and Xueguang Ma and Shengyao Zhuang and Jimmy Lin and Akari Asai and Victor Zhong},
 title = {AgentIR: Reasoning-Aware Retrieval for Deep Research Agents},
 year = {2026}
}

@Article{Jin2026LaSERIE,
 author = {Jiajie Jin and Yanzhao Zhang and Mingxin Li and Dingkun Long and Pengjun Xie and Yutao Zhu and Zhicheng Dou},
 booktitle = {arXiv.org},
 journal = {ArXiv},
 title = {LaSER: Internalizing Explicit Reasoning into Latent Space for Dense Retrieval},
 volume = {abs/2603.01425},
 year = {2026}
}

@Inproceedings{Abdallah2026AreLR,
 author = {Abdelrahman Abdallah and J. Holdcroft and M. Ali and Adam Jatowt},
 title = {Are LLM-Based Retrievers Worth Their Cost? An Empirical Study of Efficiency, Robustness, and Reasoning Overhead},
 year = {2026}
}

@Inproceedings{Xiong2026AutoResearchBenchBA,
 author = {Lei Xiong and Kun Luo and Ziyi Xia and Wenbo Zhang and Jin-Ge Yao and Zheng Liu and Jing Shao and Jianlyu Chen and Hongjin Qian and Xi Yang and Qian Yu and Hao Li and C. Yue and Xia'an Du and Yuyang Wang and Yesheng Liu and Haiyu Xu and Zhicheng Dou},
 title = {AutoResearchBench: Benchmarking AI Agents on Complex Scientific Literature Discovery},
 year = {2026}
}

@Article{Li2026DeepResearchBI,
 author = {Ruizhe Li and Mingxuan Du and Benfeng Xu and Chiwei Zhu and Xiaorui Wang and Zhendong Mao},
 booktitle = {arXiv.org},
 journal = {ArXiv},
 title = {DeepResearch Bench II: Diagnosing Deep Research Agents via Rubrics from Expert Report},
 volume = {abs/2601.08536},
 year = {2026}
}

@Article{Gupta2026DeepSearchQABT,
 author = {Nikita Gupta and Riju Chatterjee and Lukas Haas and Connie Tao and Andrew Wang and Chang Liu and Hidekazu Oiwa and E. Gribovskaya and Jan Ackermann and John Blitzer and S. Goldshtein and D. Das},
 booktitle = {arXiv.org},
 journal = {ArXiv},
 title = {DeepSearchQA: Bridging the Comprehensiveness Gap for Deep Research Agents},
 volume = {abs/2601.20975},
 year = {2026}
}

@Inproceedings{Samuel2026CoverageBenchEI,
 author = {Saron Samuel and Andrew Yates and Dawn J. Lawrie and Ian Soboroff and Trevor Adriaanse and Benjamin Van Durme and Eugene Yang},
 title = {CoverageBench: Evaluating Information Coverage across Tasks and Domains},
 year = {2026}
}

@Inproceedings{Wu2026AgentSearchBenchAB,
 author = {Bin Wu and Arastun Mammadli and Xiaoyu Zhang and Emine Yilmaz},
 title = {AgentSearchBench: A Benchmark for AI Agent Search in the Wild},
 year = {2026}
}

@inproceedings{
zhao2026sciarena,
title={SciArena: An Open Evaluation Platform for Non-Verifiable Scientific Literature-Grounded Tasks},
author={Yilun Zhao and Kaiyan Zhang and Tiansheng Hu and Sihong Wu and Ronan Le Bras and Yixin Liu and Xiangru Tang and Joseph Chee Chang and Jesse Dodge and Jonathan Bragg and Chen Zhao and Hannaneh Hajishirzi and Doug Downey and Arman Cohan},
booktitle={The Thirty-ninth Annual Conference on Neural Information Processing Systems Datasets and Benchmarks Track},
year={2026},
url={https://openreview.net/forum?id=am6RR85mnc}
}

@Article{Wang2026TrainingDR,
 author = {Benben Wang and Minghao Tang and Hengran Zhang and Jiafeng Guo and Keping Bi},
 booktitle = {arXiv.org},
 journal = {ArXiv},
 title = {Training Dense Retrievers with Multiple Positive Passages},
 volume = {abs/2602.12727},
 year = {2026}
}

@Article{Esfandiarpoor2025BeyondCL,
 author = {Reza Esfandiarpoor and George Zerveas and Ruochen Zhang and Macton Mgonzo and Carsten Eickhoff and Stephen H. Bach},
 booktitle = {Conference on Empirical Methods in Natural Language Processing},
 pages = {22860-22882},
 title = {Beyond Contrastive Learning: Synthetic Data Enables List-wise Training with Multiple Levels of Relevance},
 year = {2025}
}

@Article{Dou2025FLeWFA,
 author = {Zheng Dou and Deqing Wang and Fuzhen Zhuang and Jian Ren and Yanling Hu},
 booktitle = {arXiv.org},
 journal = {ArXiv},
 title = {FLeW: Facet-Level and Adaptive Weighted Representation Learning of Scientific Documents},
 volume = {abs/2509.07531},
 year = {2025}
}

@Article{Zhou2025ARKAR,
 author = {Jiawei Zhou and Hang Ding and Haiyun Jiang},
 booktitle = {arXiv.org},
 journal = {ArXiv},
 title = {ARK: Answer-Centric Retriever Tuning via KG-augmented Curriculum Learning},
 volume = {abs/2511.16326},
 year = {2025}
}

@Article{Moreira2025ImprovingTE,
 author = {G. Moreira and Radek Osmulski and Mengyao Xu and Ronay Ak and Benedikt Schifferer and Even Oldridge},
 booktitle = {International Conference on Information and Knowledge Management},
 journal = {Proceedings of the 34th ACM International Conference on Information and Knowledge Management},
 title = {Improving Text Embedding Models with Positive-aware Hard-negative Mining},
 year = {2025}
}

@Article{Feng2026TheWO,
 author = {Xincan Feng and Noriki Nishida and Yusuke Sakai and Yuji Matsumoto},
 booktitle = {arXiv.org},
 journal = {ArXiv},
 title = {The Wisdom of Many Queries: Complexity-Diversity Principle for Dense Retriever Training},
 volume = {abs/2602.09448},
 year = {2026}
}

@Article{Zhang2025MRMRAR,
 author = {Siyue Zhang and Yuan Gao and Xiao Zhou and Yilun Zhao and Tingyu Song and Arman Cohan and A. Luu and Chen Zhao},
 booktitle = {arXiv.org},
 journal = {ArXiv},
 title = {MRMR: A Realistic and Expert-Level Multidisciplinary Benchmark for Reasoning-Intensive Multimodal Retrieval},
 volume = {abs/2510.09510},
 year = {2025}
}

@Article{Song2025LimRankLI,
 author = {Tingyu Song and Yilun Zhao and Siyue Zhang and Chen Zhao and Arman Cohan},
 booktitle = {Conference on Empirical Methods in Natural Language Processing},
 journal = {ArXiv},
 title = {LimRank: Less is More for Reasoning-Intensive Information Reranking},
 volume = {abs/2510.23544},
 year = {2025}
}

@Article{Ding2026SciRAGAC,
 author = {Hang Ding and Yilun Zhao and Tiansheng Hu and Manasi Patwardhan and Arman Cohan},
 booktitle = {Conference of the European Chapter of the Association for Computational Linguistics},
 pages = {6440-6460},
 title = {SciRAG: Adaptive, Citation-Aware, and Outline-Guided Retrieval and Synthesis for Scientific Literature},
 year = {2026}
}

@Article{Song2025IFIRAC,
 author = {Tingyu Song and Guo Gan and Mingsheng Shang and Yilun Zhao},
 booktitle = {North American Chapter of the Association for Computational Linguistics},
 pages = {10186-10204},
 title = {IFIR: A Comprehensive Benchmark for Evaluating Instruction-Following in Expert-Domain Information Retrieval},
 year = {2025}
}

@Article{Hu2026SAGEBA,
 author = {Tiansheng Hu and Yilun Zhao and Canyu Zhang and Arman Cohan and Chen Zhao},
 booktitle = {arXiv.org},
 journal = {ArXiv},
 title = {SAGE: Benchmarking and Improving Retrieval for Deep Research Agents},
 volume = {abs/2602.05975},
 year = {2026}
}
